%% file: acl_latex.tex
\pdfoutput=1

\documentclass[11pt]{article}

\usepackage[final]{acl}

\usepackage{times}
\usepackage{latexsym}

\usepackage[T1]{fontenc}

\usepackage[utf8]{inputenc}

\usepackage{microtype}

\usepackage{inconsolata}

\usepackage{amsmath,amsfonts,bm,amsthm}
\usepackage{xcolor}
\def\myalgo{\textsc{SWEET}}
\def\wllm{\textsc{WLLM}}
\def\rdfw{\textsc{EXP-edit}}
\def\vl{\bm{l}}
\def\vp{\bm{p}}
\def\vocab{\mathcal{V}}
\def\threshold{\tau}
\usepackage{graphicx}
\usepackage{booktabs}
\usepackage{multirow}
\usepackage{hyperref}
\usepackage{listings}
\usepackage{bbold}
\lstdefinestyle{base}{
  language=Python,
  emptylines=1,
  breaklines=true,
  moredelim=**[is][\color{red}]{@}{@},
}

\newcommand{\cam}[1]{{#1}}

\newcommand{\update}[1]{{#1}}

\title{Who Wrote this Code? Watermarking for Code Generation}

\author{
\quad Taehyun Lee$^{\star, 1}$
\quad Seokhee Hong$^{\star, 1, 3}$
\quad \textbf{Jaewoo Ahn}$^{1}$
\quad \textbf{Ilgee Hong}$^{1, 4, \sharp}$ \\ 
\quad \textbf{Hwaran Lee}$^{2}$
\quad \textbf{Sangdoo Yun}$^{1, 2}$
\quad \textbf{Jamin Shin}$^{\dagger, 2,\natural}$
\quad \textbf{Gunhee Kim}$^{\dagger, 1}$\\
$^1$Seoul National University\quad $^2$NAVER AI Lab \\
\quad $^3$LG AI Research \quad $^4$Georgia Institute of Technology\\
\texttt{\small \{taehyun.lee, seokhee.hong, jaewoo.ahn\}@vision.snu.ac.kr} \\
\texttt{\small ihong39@gatech.edu,} \texttt{\small \{hwaran.lee, sangdoo.yun\}@navercorp.com}\\
\texttt{\small jamin.shin@outlook.com, gunhee@snu.ac.kr} \\
}

\newcommand{\correspondingfootnote}{
    \let\oldthefootnote=\thefootnote
    \renewcommand{\thefootnote}{}
    \footnotetext{$\star$ Authors equally contributed.}
    \footnotetext{$\sharp$ Work done during Ilgee's internship at SNU VL Lab.}
    \footnotetext{$\natural$ Jamin Shin is currently affiliated with Trillion Labs, but the work was done while he was at NAVER. }
    \footnotetext{$\dagger$ Corresponding authors.}
    \let\thefootnote=\oldthefootnote
}
\newtheorem{lemma}{Lemma}[section] 
\newtheorem{theorem}{Theorem}
\begin{document}
\maketitle

\input{sections/00_abstract}

\correspondingfootnote

\input{sections/01_intro}
\input{sections/02_related}
\input{sections/03_method}
\input{sections/04_experiment}
\input{sections/05_results}
\input{sections/06_analysis}
\input{sections/07_conclusion}
\input{sections/08_limitations}

\input{sections/ethical}

\input{sections/ack}

\bibliography{custom}

\input{sections/appendix}

\end{document}

%% file: sections/00_abstract.tex
\begin{abstract}
Since the remarkable generation performance of large language models raised ethical and legal concerns, approaches to detect machine-generated text by embedding watermarks are being developed.
However, we discover that the existing works fail to function appropriately in code generation tasks due to the task's nature of having low entropy.
Extending a logit-modifying watermark method, we propose \textbf{S}elective \textbf{W}at\textbf{E}rmarking via \textbf{E}ntropy \textbf{T}hresholding (\myalgo), which enhances detection ability and mitigates code quality degeneration by removing low-entropy segments at generating and detecting watermarks.
Our experiments show that \myalgo~significantly improves code quality preservation while outperforming all baselines, including post-hoc detection methods, in detecting machine-generated code text.
Our code is available in
\url{https://github.com/hongcheki/sweet-watermark}.
\end{abstract}

%% file: sections/01_intro.tex
\input{resources_acl/fig_introduce}

\section{Introduction}\label{sec:intro}
In understanding and generating software programs, large language models have rapidly advanced towards expert-like proficiency~\citep{Chen2021HumanEval,Luo2023WizardCoder,Li2023StarCoder,Nijkamp2023CodeGen,Zheng2023Codegeex,gunasekar2023textbooks,touvron2023llama2,OpenAI2023gpt4}.
This breakthrough in the automation of the coding process improves the productivity and efficiency of software engineer and lowers the barriers to creating programs for non-experts~\citep{Vaithilingam2022Expectation}.

However, this advance comes with significant legal, ethical, and security concerns, including code licensing issues, code plagiarism, code vulnerability, and malware generation \cite{He2023Controlling,Sandoval2023Lost,Pearce2022Asleep,Carlini2021Extracting,Mirsky2022threat,hazell2023large}. For example, there is an ongoing class-action copyright lawsuit between a group of individuals and Microsoft, GitHub, and OpenAI, arising from allegations of unlawful utilization and reproduction of the source code\footnote{\href{https://drewdevault.com/2022/06/23/Copilot-GPL-washing.html}{Code plagiarism}}\footnote{\href{https://www.reuters.com/legal/litigation/openai-microsoft-want-court-toss-lawsuit-accusing-them-abusing-open-source-code-2023-01-27/}{Code licensing issue}}. Furthermore, shortly after the launch of ChatGPT, numerous malicious actors on the Dark Web were observed sharing machine-generated malware and spear phishing tutorials\footnote{\href{https://www.recordedfuture.com/i-chatbot}{Malware generation}}. Therefore, the development of reliable tools for detecting machine-generated code is a very timely matter and is of utmost importance for fairly deploying LLMs with coding capabilities.

Despite the need for immediate treatment of the machine-generated code detection problem, few efforts have been made to address it. Instead, many works still prioritize a detection problem on normal text~\citep{Solaiman2019Release,Ippolito2020Automatic,Guo2023How, tian2023gptzero, OpenAI2023classifier,Yu2023GPT,Gehrmann2019GLTR,Mitchell2023Detectgpt,Yang2023DNA}.
While these \textit{post-hoc} detection methods (i.e., no control during the text generation) have demonstrated powerful performance in the many domain of natural language tasks, their application to programming language remains unexplored.

\update{Contrary to the post-hoc detection methods, another line of research for detecting machine-generated text has gained attention: \textit{Watermarking-based} methods, which embed a hidden signal within the generated text~\citep{Kirchenbauer2023watermark,kirchenbauer2023reliability,kuditipudi2023robust,wang2023towards}.}
\update{For example, a method proposed in \citet{Kirchenbauer2023watermark}} -- which we refer to as \wllm~(Watermarking for Large Language Models) -- randomly divides the entire vocabulary into two groups \update{(i.e., the green list and the red list) at each generation step and enhance the probability of green list tokens to be sampled. By adding scalar values to the logits of a green list tokens,} the model favors generating tokens from the green list rather than the red one.
To detect the watermark in a text, we count the number of green tokens and check whether this number is statistically significant (through hypothesis testing) to conclude whether the model output is generated without knowledge of the green-red rule. 

While both watermarking-based methods and post-hoc detection methods work well in many language generation tasks, we observe that these performances do not transfer well to code generation tasks, for example, in Figure~\ref{fig:introduce}.
\update{In other words, it is much more challenging to embed watermarks in a detectable way without impairing the code functionality.}
We attribute this to the nature of extremely low entropy\footnote{We calculate entropy over the probability of the next token prediction.
Please refer to Eq. \ref{eq:entropy} for details.} of code generation. 
If watermarking is applied strongly, it can severely degrade the quality of the model output, which is particularly critical in code generation, as a single violation of a rule can break the entire code (see ``strong watermark'' in Figure~\ref{fig:introduce}). On the other hand, if watermarking is applied too weakly, \update{the low entropy hinders properly embedding watermarks and results in insufficient green tokens appearing,} leading to increased difficulty in detection (see ``weak watermark'' in Figure~\ref{fig:introduce}). These failures are not significant in plain text generation because the relatively higher entropy allows for more flexibility in candidate selections for watermarking.

To address these failure modes, we extend the WLLM and propose \textbf{S}elective \textbf{W}at\textbf{E}rmarking via \textbf{E}ntropy \textbf{T}hresholding ($\myalgo$) for Code LLMs (and LLMs). Instead of applying the green-red rule to every single token during generation, we only apply the rule to tokens with \textit{high enough entropy} given a threshold. That is, we do not apply the green-red rule to the important tokens for making functional code, while making sure there are enough green list tokens to make a detectable watermark for less important tokens, hence, directly addressing each of the above failure modes.
\update{In code generation tasks, our method outperforms all baselines, including post-hoc detection methods, in detecting machine-generated code while achieving less code quality degradation than~\wllm. Furthermore, through various analyses, we demonstrate that our method operates well even without prompts or with a small surrogate model, indicating its robust performance under practical settings.}

\update{Our contributions are as follows:}
\begin{itemize}
    \item We are the first to empirically explore the breakdown of existing watermarking and \update{post-hoc} detection methods in the code domain.
    \item We propose a simple yet effective method called $\myalgo$, which improves $\wllm$ \cite{Kirchenbauer2023watermark} and achieves significantly higher \update{performance in machine-generated code detection while preserving code quality more than $\wllm$.}
    \item \update{We have demonstrated the practical applicability and predominance of our method even in real-world settings, i.e., 1) without prompts, 2) utilizing a smaller model as a detector, or 3) under paraphrasing attacks.}
\end{itemize}

%% file: resources_acl/fig_introduce.tex
\begin{figure}[hbt!]
\begin{center}
\includegraphics[width=\linewidth]{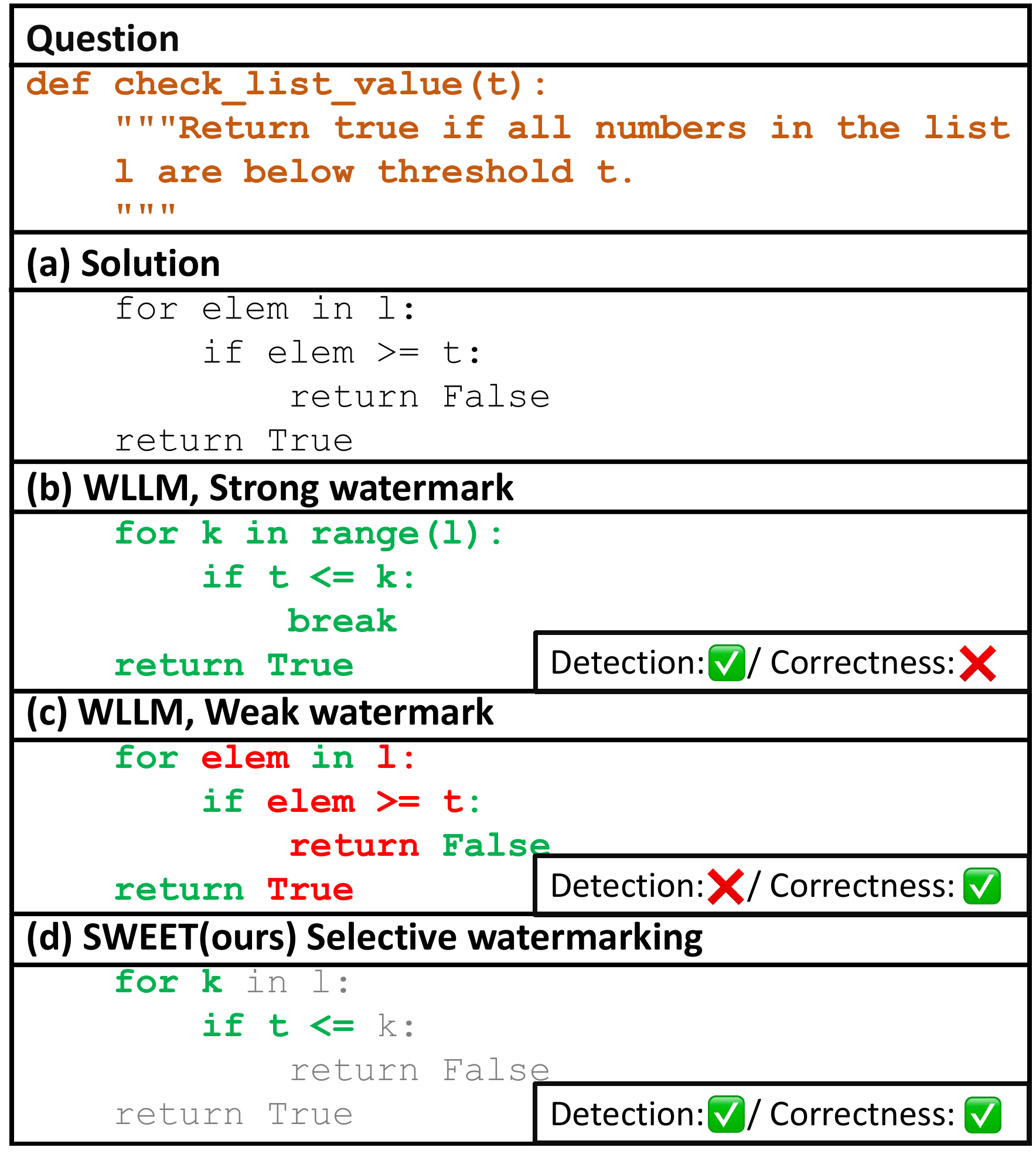}  
\end{center}
\caption{
Illustrated comparison of $\wllm$ \citep{Kirchenbauer2023watermark} and $\myalgo$ (ours). Note that this example is a short hypothetical explanatory example. LLMs can generate working source code (a) without a watermark. Strong watermark (b) or weak watermark (c) may result in detection or correctness failure, but (d) selective watermarking may avoid both failures. 
}
\label{fig:introduce}
\end{figure}

%% file: sections/02_related.tex
\section{Related Work}

\textbf{Software Watermarking}
Software watermarking is the research field where a secret signal is embedded in the code without affecting its performance, to prevent software piracy.
Static watermarking~\citep{Hamilton2011survey, Li2010Design, Myles2005evaluation} imprints watermarks typically through code replacement and reordering.
On the other hands, dynamic watermarking~\citep{Wang2018Exception, Ma2019Xmark} injects watermarks during the compiling or executing stage of a program. For a detailed survey, please refer to~\citet{Dey2019Software}.

\update{Watermarking code text generated from a LLM is closer to static watermarking. For example, \citet{li2023protecting} proposes a method employing the replacement of synonymous code.
However, since this method heavily relies on language-specific rules, a malicious user knowing these rules could reverse the watermarking.}

\textbf{LLM Text Watermarking}
The majority of watermarking methods for texts \update{from LLMs} are based on the modification of the original text via a predefined set of rules~\citep{Atallah2001Natural, Atallah2003Natural, Kim2003text, Topkara2006hiding, Jalil2009review, Meral2009Natural,he2022protecting,he2022cater} or another language model, such as transformer-based networks.~\citep{Abdelnabi2021Adversarial, Yang2022Tracing, Yoo2023Robust}.

\update{Recently, a line of work embeds watermarks into tokens during the sampling process of LLMs~\citep{liu2024survey}. They embed watermarks within LLM-generated texts by either motifying logits from the LLM~\citep{Kirchenbauer2023watermark,kirchenbauer2023reliability,liu2023unforgeable,takezawa2023necessary,hu2023unbiased} or manipulating the sampling procedure~\citep{christ2023undetectable,kuditipudi2023robust}.}
\update{Moreover, some recent works focus on the robustness of watermarks against attacks to remove watermarks~\citep{zhao2023provable,liu2023semantic,ren2023robust}. Lastly, \citet{gu2023learnability} investigates the learnability of watermarks in the distillation process from teacher to student model.}




\update{However, these watermark methods exhibit vulnerability in their watermark detection performance under low entropy situations~\citep{Kirchenbauer2023watermark, kuditipudi2023robust}, and a limited number of studies, such as CTWL~\citep{wang2023towards}, try to handle it. We directly address the degradation of watermark detection performance in low entropy situations and demonstrate our method's efficacy in low entropy tasks, such as code generation.}

\textbf{Post-hoc Detection}
\update{Post-hoc detection methods aim to differentiate between human-authored and machine-generated text without embedding any signal during generation.}
One line of work leverages perplexity-based features like GPTZero~\citep{tian2023gptzero}, Sniffer~\citep{li2023sniffer}, and LLMDet~\citep{Wu2023LLMDet}.
Another line of work uses pre-trained LM, such as RoBERTa~\citep{Liu2019Roberta}, and fine-tunes it as a classifier to identify the source of text~\citep{Solaiman2019Release, Ippolito2020Automatic,OpenAI2023classifier,Guo2023How,Yu2023GPT,mitrović2023chatgpt}.
Meanwhile, some recent works tackle the detection problem without additional training procedures, such as GLTR~\citep{Gehrmann2019GLTR}, DetectGPT~\citep{Mitchell2023Detectgpt}, and DNA-GPT~\citep{Yang2023DNA}.
\update{However, post-hoc detection methods remain challenging. For example, while the GPTZero~\citep{tian2023gptzero} is still in service, OpenAI's AI text classifier~\citep{OpenAI2023classifier} was discontinued after six months due to low accuracy rates. Furthermore, we have demonstrated that post-hoc detection methods failed to detect machine-generated code, with low entropy.}

%% file: sections/03_method.tex
\input{resources_acl/fig_real_examples}

\section{Method}

We propose a new watermarking method, $\myalgo$, that selectively watermarks tokens only with high enough entropy. 

\subsection{Motivation}\label{ssec:motivation}
Although the previous watermarking method WLLM~\citep{Kirchenbauer2023watermark} can be applied to any domain of LLM-generated text\footnote{Please refer to Appendix~\ref{appendix:preliminaries} for the details of $\wllm$.}, it incurs two critical problems during embedding and detecting watermarks in code generation, attributed to a dilemma regarding watermark strength.

\textbf{Watermarking causes performance degradation.}
\update{There are only a few different ways of expressing the same meaning in a programming language, and just one wrong token can be attributed to undesirable outputs.}
\update{If watermarks are embedded strongly, as $\wllm$ randomly divides the vocabulary into green and red lists without leveraging any information about the context, promoting the logits of only green list tokens must heighten the chance of generating the wrong token.}
For example, in Figure~\ref{fig:real_examples} (a), after ``return'' token in the second row, the next token with the highest logit is ``sum'', which is also part of the canonical solution.
However, $\wllm$ puts ``sum'' into the red list while putting ``mean'' into the green list. Hence, the sampled token was ``mean'', resulting in a syntax error.

\textbf{Low Entropy Sequences Avoid Being Watermarked.}
Another critical issue is when watermark strength is too weak to embed watermarks into a text with low entropy. If a red list token has a too high logit value to be inevitably generated, it hinders watermark detection.
For example, in Figure~\ref{fig:real_examples} (a), tokens with white backgrounds representing low entropy have few green tokens.
This becomes much more fatal in code generation tasks where outcomes are relatively shorter than the plain text, such as asking only a code block of a function\footnote{The average token length of human-written solution codes in HumanEval, MBPP, and DS-1000 datasets is only 57.}.
The $\wllm$ detection method is based on a statistical test, which involves counting the number of green list tokens in the entire length. However \update{detecting watermarks based on a statistical test deteriorates if the length is short.}\footnote{\update{We measured detectability according to the length of generated texts and observed that $\wllm$ performs relatively poorly while $\myalgo$ is robust in detecting watermarks within short texts. For more details, please refer to Appendix~\ref{appedix:detectability@T}.}}

\subsection{The SWEET Method}

SWEET can mitigate this dilemma regarding the watermark strength by distinguishing watermark-applicable tokens\update{, meaning we embed and detect watermarks only within tokens with high entropy.}

\textbf{Generation.} The generation step of our method is in Algorithm~\ref{alg:alg1}. \update{Given a tokenized prompt $\bm{x}=\{x_0,\dots,x_{M-1}\}$ and already generated tokens $\bm{y}_{[:t]}=\{y_0,\dots,y_{t-1}\}$, a model calculates an entropy value ($H_t$) of the probability distribution for $y_t$. We then only apply the watermarking when $H_t$ is higher than the threshold, $\threshold$.}
We randomly bin a vocabulary by green and red with a fixed green token ratio $\gamma$. If a token is selected to be watermarked, we add a constant $\delta$ to green tokens' logits, aiming to promote the sampling of the green tokens.
\update{By limiting the promotion of green tokens only to tokens with high entropy, we prevent the model's logit distribution changes for tokens where the model has confidence (and, therefore, low entropy), resulting in preserving code quality.}

\textbf{Detection.} We outline our detection process in Algorithm~\ref{alg:alg2}. Given a token sequence $\bm{y}=\{y_0,\dots,y_{N-1}\}$, our task is to detect watermarks within $\bm{y}$; therefore, determine whether it is generated from the specific language model.
Like in the generation phase, we compute the entropy values $H_t$ for each $y_t$. Let $N^h$ denote the number of tokens that have an entropy value $H_{t}$ higher than the threshold $\threshold$, and let $N^h_G$ denote the number of green tokens among in $N^h$. Finally, with the green list ratio among entire vocabulary $\gamma$ used in the generation step, we compute a $z$-score under the null hypothesis where the text is not watermarked by
\begin{equation}\label{eq:z-stat}
z=\frac{N^h_G-\gamma N^h}{\sqrt{N^h\gamma(1-\gamma)}}
\end{equation}
We can say the text is watermarked more confidently as $z$-score goes higher. We set $z_{\text{threshold}}$ as a cut-off score. 
If $z>z_{\text{threshold}}$ holds, we decide that the watermark is embedded in $\bm{y}$ and thus generated by the LLM.
\update{The effect of the entropy threshold in the detection phase is described in the following section.}

\subsection{\update{Effect of Entropy Thresholding}}
This section shows that selective watermark detection based on the entropy threshold improves the detectability.

Theorem~\ref{theorem:1} implies that we can ensure a higher lower bound of $z$-score by the $\myalgo$ detection method than $\wllm$. Recalling Sec~\ref{ssec:motivation}, this is achieved by ignoring tokens with low entropy, leading to increases in the ratio of green tokens within the text and detectability.

For the sake of theoretical analysis, we use \textit{spike entropy} (Eq.~\ref{eq:spike_entropy}), which is a variant of entropy defined in~\citet{Kirchenbauer2023watermark}. In practice, we use the entropy in Eq.~\ref{eq:entropy}.

\begin{theorem}\label{theorem:1}
Consider a token sequence $\bm{y}=\{y_0,\dots,y_{N-1}\}$ generated by a watermarked code LLM. $(S_0,\dots,S_{N-1})$ is a sequence of corresponding spike entropy, in which the modulus is $\frac{(1-\gamma)(e^\delta-1)}{1+(e^\delta-1)\gamma}$. Let $\threshold$ be an entropy threshold, $N^{l}$ and $N^{h}$ be the number of tokens whose spike entropy is lower or higher than the threshold.

If the following assumption regarding the ratio of low entropy tokens holds
\begin{equation}\label{eq:assumption}
\frac{N^{l}}{N} \le 1-(\frac{\alpha\overline{S}-1}{\alpha\overline{S^h}-1})^2\nonumber
\end{equation}
then there is a lower bound of $z$-score that is always higher when the entropy threshold is applied,
where $\alpha=\frac{e^\delta}{1+(e^\delta-1)\gamma}$, $\overline{S}=\Sigma_{t=1}^N S_t/N$, and $\overline{S^h}=\Sigma_{t=1}^N S_t \times \mathbb{1} (S_t \ge \threshold)/N^h$.
\end{theorem}

\textit{Remark.} The assumption means choosing an entropy threshold that does not ignore too many tokens ($N^l$) is important. 






%% file: resources_acl/fig_real_examples.tex
\begin{figure*}[hbt!] 
\begin{center}
\includegraphics[width=\textwidth]{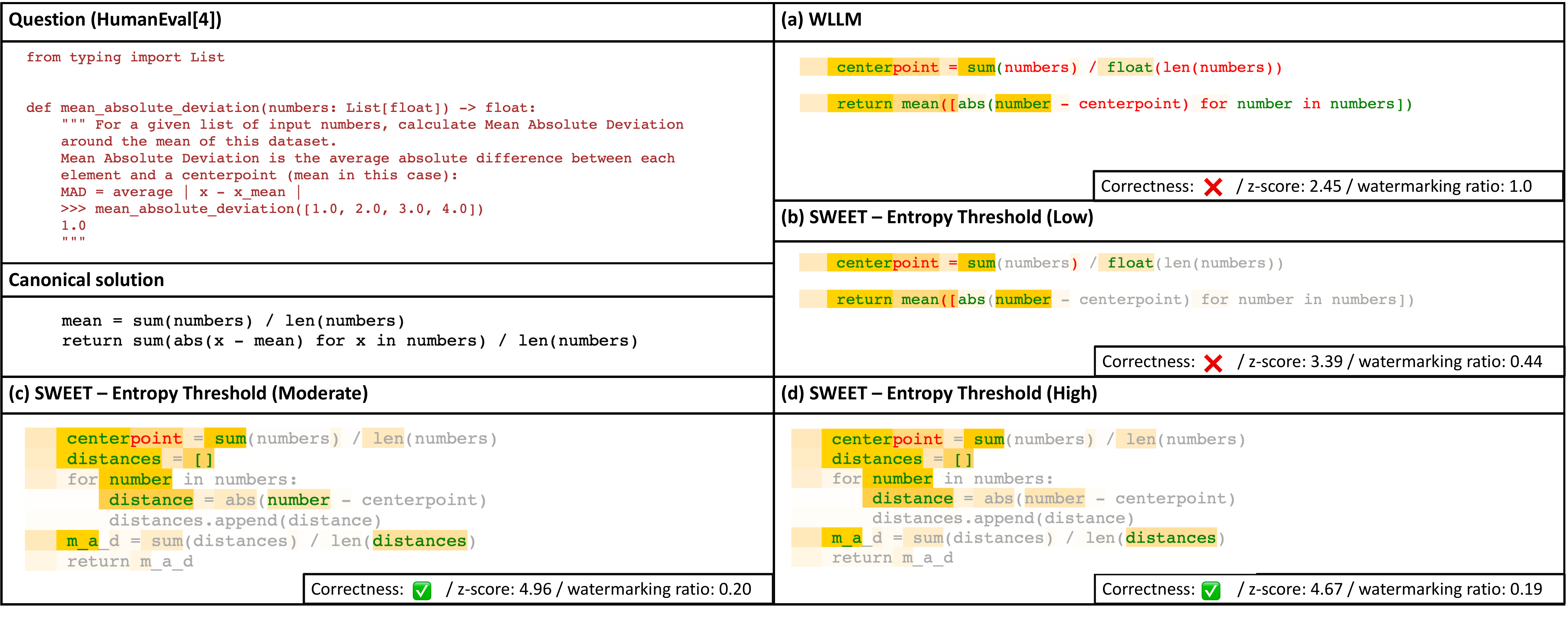}  
\end{center}
\caption{
A \update{real} example of HumanEval/4  for comparing between (a) WLLM and (b)--(d) our SWEET with different thresholds. Text colors annotate whether tokens are in the green or red list. Gray tokens have entropy smaller than the threshold and are not watermarked. The intensity of the yellow background color visualizes the entropy value. \update{(a) While~\wllm~produces an incorrect code and less detectable watermarks with a few green tokens (low z-score), (b)-(d)~\myalgo~improves both code quality and z-score by selectively embedding and detecting watermarks using an entropy threshold. Interestingly, (c) the z-score peaks with a moderate threshold, and (d) as the threshold increases, the z-score declines due to the decrease in the watermarking ratio.}
}
\label{fig:real_examples}
\end{figure*}


%% file: sections/04_experiment.tex
\input{resources_acl/table_main}

\section{Experiments}\label{sec:experiment}

We conduct a series of experiments to evaluate the effectiveness of our watermarking method in code generation for two aspects: (i) quality preserving ability and (ii) detection strength. Our base model is StarCoder~\citep{Li2023StarCoder}, which is an open-source LLM specifically for code generation. We also conduct experiments on one of the general-purpose LLM, LLaMA\update{2}~\citep{touvron2023llama2} (see the results in Appendix~\ref{appendix:pareto}).
\subsection{Tasks and Metrics}\label{subsec:tasks_and_metrics}
We select \update{three} Python code generation tasks, HumanEval \cite{Chen2021HumanEval}, MBPP \cite{austin2021MBPP}\update{, and DS-1000 \cite{lai2023ds1000}}, as our \cam{main} testbeds. These tasks contain Python programming problems, test cases, and human-written canonical answers. Language models are prompted with programming problems and expected to generate the correct code that can pass the test cases. \cam{To evaluate our approach's performance in more diverse software development contexts, such as other languages or other code generation scopes, we also include two more datasets: HumanEvalPack~\citep{muennighoff2024octopack} and ClassEval~\citep{du2023classeval}. Please refer to Appendix~\ref{appendix:humanevalpack_classeval} for implementation details of these benchmarks.}

To evaluate the functional quality of generated source code, we use pass@k~\citep{Chen2021HumanEval} by generating $n (>k)$ outputs for each programming problems. This metric estimates the percentage of code generated correctly-performing.
For the detection ability, we use AUROC (i.e., Area Under ROC) value as a main metric. We also report the true positive rate (TPR; correctly detecting LLM-generated code as LLM-generated) when the false positive rate (FPR; falsely detecting human-written code as LLM-generated) is confined to be lower than 5\%. This is to observe the detection ratio of a practical setting, where high false positive is more undesirable than false negative.

\subsection{Baselines}\label{subsec:baselines} 
We compare $\myalgo$ with machine-generated text detection baselines. \textit{Post-hoc} detection baselines do not need any modification during generation so that they never impair the quality of the model output. \textsc{logp(x)},  \textsc{LogRank}~\citep{Gehrmann2019GLTR}, and \textsc{DetectGPT}~\citep{Mitchell2023Detectgpt} are zero-shot detection methods that need no labeled datasets. \textsc{GPTZero}~\citep{tian2023gptzero} and \textsc{OpenAI Classifier}~\citep{Solaiman2019Release} are trained classifiers.
\update{For \textit{Watermarking-based} methods, we have included two baselines: $\wllm$~\cite{Kirchenbauer2023watermark} and $\rdfw$~\cite{kuditipudi2023robust}.}
\update{To embed a watermark, methods that distort the model's sampling distribution, such as $\wllm$ or ours,} tend to have better detection ability, but degradation of text quality may arise. On the other hand, $\rdfw$ is expected to cause no degradation in text quality as they do not distort the sampling distribution of the model. \footnote{When evaluating code generation performance through pass@1, a low temperature was applied to all models. However, the spiky distribution resulting from the low temperature hindered $\rdfw$ from adequately embedding watermarking. Therefore, we have also included $\rdfw$ baseline with a high entropy by setting temperature=1.0 and top-p=1.0.} More details of implementation are in Appendix~\ref{appendix:Implementation Details}.

%% file: resources_acl/table_main.tex
\begin{table*}[th]
\centering
\resizebox{\textwidth}{!}{%
\begin{tabular}{@{}crlcccccccccccccc@{}}
\toprule
\multicolumn{2}{c}{\multirow{2}{*}{\textbf{Method}}} &
   &
  \multicolumn{4}{c}{\textbf{\textsc{HumanEval}}} &
  \textbf{} &
  \multicolumn{4}{c}{\textbf{\textsc{MBPP}}} &
   &
  \multicolumn{4}{c}{\update{\textbf{\textsc{DS-1000}}}} 
  \\ \cmidrule(lr){4-7} \cmidrule(lr){9-12} \cmidrule(lr){14-17}
\multicolumn{2}{c}{}                                                  &  & \textsc{pass@1} & AUROC & TPR & FPR &  & \textsc{pass@1} & AUROC & TPR & FPR &  & \textsc{pass@1} & AUROC & TPR & FPR \\ \midrule
\multicolumn{2}{c}{\textbf{Non-watermarked}}                          &  & \textbf{33.4} & - & - & - &  & \textbf{37.8} & - & - & - &  & \textbf{26.3} & - & - & -  \\ 
\multicolumn{2}{c}{\update{\textbf{Non-watermarked (w/ high entropy)}}}                          &  & 18.3 & - & - & - &  & 21.4 & - & - & - &  & 12.7 & - & - & -  \\ \midrule
\multicolumn{17}{l}{\textbf{\textit{Post-hoc}}}     \\
\multirow{6}{*}{\textbf{}} & \textsc{log p(x)}               &  & \multirow{6}{*}{\textbf{33.4}} & 0.533 & 0.113 & < 0.05 &  & \multirow{6}{*}{\textbf{37.8}} & 
0.525 & 0.054 & < 0.05 &  & \multirow{6}{*}{\textbf{26.3}} & 0.566 & 0.100 & < 0.05 \\
                                    & \textsc{LogRank}                &  &          & 0.553 & 0.127 & < 0.05 &  &         & 0.527 & 0.052 & < 0.05 &  &         & 0.562 & 0.105 & < 0.05 \\
                                    & \textsc{DetectGPT (T5-3B)}      &  &          & 0.549 & 0.092 & < 0.05 &  &         & 0.531 & 0.040 & < 0.05 &  &         & 0.433 & 0.070 & < 0.05 \\
                                    & \textsc{DetectGPT} &  &          & 0.533 & 0.165 & < 0.05 &  &         & 0.565 & 0.158 & < 0.05 &  &         & 0.606 & 0.113 & < 0.05 \\ 
                                    & \textsc{GPTZero} &  &          & 0.521 & 0.122 & < 0.05 &  &         & 0.449 & 0.026 & < 0.05 &  &         & 0.539 & 0.063 & < 0.05 \\ 
                                    & \textsc{OpenAI Classifier} &  &          & 0.518 & 0.053 & < 0.05 &  &         & 0.500 & 0.036 & < 0.05 &  &         & 0.524 & 0.075 & < 0.05 \\ \midrule
\multicolumn{17}{l}{\textbf{\textit{Watermarking}}}     \\
\multirow{6}{*}{\textbf{}} & \update{\rdfw} & & \textbf{33.6} & 0.489 & 0.085 & < 0.05 & & \textbf{37.5} & 0.536 & 0.044 & < 0.05 & & \textbf{26.2} & 0.546 & 0.066 & < 0.05 \\
                                    & \update{\rdfw~(w/ high entropy)} & & 19.3 & 0.733 & 0.427 & < 0.05 & & 22.7 & 0.744 & 0.33 & < 0.05 & & 12.7 & 0.743 & 0.378 & < 0.05 \\ \cmidrule(lr){4-17}
                                    & \wllm~($\Delta$\textsc{pass@1}~$\sim-10\%$)$^{\star}$          &  & 29.6 & 0.822 & 0.402 & < 0.05 &  & 34.5 & 0.718 & 0.178 & < 0.05 &  & 23.9 & 0.627 & 0.152 & < 0.05 \\
                                    & \myalgo~($\Delta$\textsc{pass@1}~$\sim-10\%$)$^{\star}$     &  & 32.6 & \textbf{0.943} & \textbf{0.835} & < 0.05 &  & 33.8 & \textbf{0.873} & \textbf{0.590} & < 0.05 &  & 23.7 & \textbf{0.815} & \textbf{0.384} & < 0.05  \\ \cmidrule(lr){4-17}
                                    & \wllm~(AUROC$\geq0.9$)$^{\dagger}$          &  & 25.3 & 0.904 & 0.652 & < 0.05 &  & 24.2 & 0.930 & 0.718 & < 0.05 &  & 8.6 & 0.944 & 0.793 & < 0.05 \\
                                    & \myalgo~(AUROC$\geq0.9$)$^{\dagger}$     &  & 32.6 & 0.943 & 0.835 & < 0.05 &  & 33.2 & 0.906 & 0.548 & < 0.05 &  & 18.8 & 0.924 & 0.649 & < 0.05 \\ \bottomrule
\end{tabular}%

}
\caption{\textbf{Main results} of code generation performance and detection ability. Since calibration on watermarking strength leads to trade-offs between code generation quality and detection ability, we present two results for \update{$\wllm$ and $\myalgo$}. $^{\star}$ for the best detection score (i.e., AUROC and TPR) while allowing a code generation quality decrease of $\sim$10\% compared to Non-watermarked, and $^{\dagger}$ for the best code generation quality (\textsc{pass@1}) among AUROC \update{$\geq$} 0.9. The selected points are shown in Figure~\ref{fig:pareto_frontier_main}. \update{We add $\rdfw$ and a Non-watermarked baseline with a high entropy setting (i.e., temperature=1.0 and top-p=1.0).
}
}
\label{tab:table_main}
\end{table*}


%% file: sections/05_results.tex
\section{Results}\label{sec:result}

\input{resources_acl/resources_appendix/tab_main_2_w_classeval.tex}
\input{resources_acl/fig_pareto_frontier_main}

\subsection{Main Results}
\label{sec:main_results}

Table~\ref{tab:table_main} presents results from all baselines and our approach.
In \update{$\wllm$ and $\myalgo$}, there is a clear trade-off between detection and code generation ability depending on the watermarking strength. Therefore, we measure the maximum scores of one domain while setting a lower bound for the scores of other domain. Specifically, to measure AUROC scores, we find the best AUROC scores around 90\% of the pass@1 performance of the non-watermarked base model. On the other hand, for measuring pass@1, we select from those with an AUROC of 0.9 or higher.

\textbf{Detection Performance}.
Table~\ref{tab:table_main} shows that overall, our $\myalgo$ method outperforms all baselines in detecting machine-generated code with a price of 10\% degradation of code functionality. \update{Both in the MBPP and DS-1000 datasets, $\myalgo$ achieves AUROC of 0.873 and 0.815, respectively, whereas none of the baselines exceeded 0.8. 
$\myalgo$ even achieves an AUROC of 0.943 in HumanEval with a 2.4\% degradation of code functionality.
However, when only near 10\% degradation of code functionality is allowed, $\wllm$ shows lower detection performance than our method. In the case of the distortion-free watermarking method, due to the lower entropy of the code generation task, $\rdfw$ fails to achieve an AUROC score exceeding 0.6 in all cases, and even $\rdfw$ with high entropy setting could not outperform our methods with regard of the detection performance.
While all post-hoc detection baselines preserve code functionality as they do not modify generated code, none of them achieve an AUROC score above 0.6.\footnote{\update{We defer a more in-depth discussion about the breakdown of Post-hoc methods to Appendix \ref{appendix:breakdown}.}}}

\textbf{Code Quality Preservation}.
\update{In the last two rows of Table~\ref{tab:table_main}, despite the inevitable text quality degradation caused by $\wllm$ and $\myalgo$,} our $\myalgo$ method preserves code functionality much more while maintaining the high detection ability of AUROC $ > 0.9$ when compared to $\wllm$. Specifically, pass@1 of $\wllm$ for HumanEval decreases from 33.4 to 25.3, a 24.3\% loss in the code execution pass rate. Similarly, for the MBPP \update{and the DS-1000} dataset, the drops in performances are 36.0\% \update{and 67.3\%, respectively}. On the other hand, our approach loses only 2.4\% (HumanEval), 12.2\% (MBPP), and 28.5\% (DS-1000), respectively, which are significantly less than those of WLLM.

\textbf{C++/Java/Class-level Code Generation}.
\cam{Table~\ref{tab:table_main_2_w_classeval} presents results on other programming languages (C++ and Java) and another code generation scope (i.e., class-level). While preserving code functionality much more than~\wllm, \myalgo~shows the highest detection performance except in the Java environment, where the TPR score of \wllm~is higher than that of \myalgo. The results demonstrate that the efficacy of our methodology is not limited to certain types of programming languages or software development environments. For more analysis of the results, please refer to Appendix~\ref{appendix:humanevalpack_classeval}.}


\subsection{Comparison of Pareto Frontiers between~\myalgo~and~\wllm}

\update{In the cases of $\myalgo$ and $\wllm$,} watermarking strength and spans can vary depending on the ratio of the green list tokens $\gamma$ and the logit increase value $\delta$. To demonstrate that $\myalgo$ consistently outperforms the baseline $\wllm$ regardless of the values of $\gamma$ and $\delta$, we draw Pareto frontier curves with axes pass@1 and AUROC in Figure~\ref{fig:pareto_frontier_main}. We observe that the Pareto frontiers of $\myalgo$ are ahead of those of $\wllm$ in all \update{three tasks. Moreover, as presented in Figure~\ref{fig:pareto_frontier_appendix}, whatever value our approach chooses for the entropy threshold, $\myalgo$ outperforms the baseline in all configurations.} This indicates that in a wide range of hyperparameter settings, our $\myalgo$ model can generate better results in terms of detection and code generation ability. Full results and different settings are in Appendix~\ref{appendix:pareto}.

%% file: resources_acl/resources_appendix/tab_main_2_w_classeval.tex
\begin{table*}[th]
\centering
\resizebox{\textwidth}{!}{%
\begin{tabular}{@{}crlcccccccccccccc@{}}
\toprule
\multicolumn{2}{c}{\multirow{2}{*}{\textbf{Method}}} &
   &
  \multicolumn{4}{c}{\textbf{\textsc{HumanEvalPack - C++}}} &
  \textbf{} &
  \multicolumn{4}{c}{\textbf{\textsc{HumanEvalPack - Java}}} &
   &
  \multicolumn{4}{c}{\update{\textbf{\textsc{ClassEval}}}} 
  \\ \cmidrule(lr){4-7} \cmidrule(lr){9-12} \cmidrule(lr){14-17}
\multicolumn{2}{c}{}                                                  &  & \textsc{pass@1} & AUROC & TPR & FPR &  & \textsc{pass@1} & AUROC & TPR & FPR &  & \textsc{pass@5} & AUROC & TPR & FPR \\ \midrule
\multicolumn{2}{c}{\textbf{Non-watermarked}}                          &  & \textbf{29.4} & - & - & - &  & \textbf{31.5} & - & - & - &  & 14.0 & - & - & -  \\ 
\multicolumn{2}{c}{\update{\textbf{Non-watermarked (w/ high entropy)}}}                          &  & 18.2 & - & - & - &  & 13.9 & - & - & - &  & 19.0 & - & - & -  \\ \midrule
\multicolumn{17}{l}{\textbf{\textit{Post-hoc}}}     \\
\multirow{6}{*}{\textbf{}} & \textsc{log p(x)}               &  & \multirow{6}{*}{\textbf{29.4}} & 0.656 & 0.160 & < 0.05 &  & \multirow{6}{*}{\textbf{31.5}} & 
0.635 & 0.127 & < 0.05 &  & \multirow{6}{*}{14.0} & 0.847 & 0.320 & < 0.05 \\
                                    & \textsc{LogRank}                &  &          & 0.658 & 0.187 & < 0.05 &  &         & 0.654 & 0.240 & < 0.05 &  &         & 0.821 & 0.260 & < 0.05 \\
                                    & \textsc{DetectGPT (T5-3B)}      &  &          & 0.646 & 0.079 & < 0.05 &  &         & 0.699 & 0.273 & < 0.05 &  &         & 0.610 & 0.140 & < 0.05 \\
                                    & \textsc{DetectGPT} &  &          & 0.525 & 0.079 & < 0.05 &  &         & 0.650 & 0.116 & < 0.05 &  &         & 0.749 & 0.210 & < 0.05 \\ 
                                    & \textsc{GPTZero} &  &          & 0.486 & 0.073 & < 0.05 &  &         & 0.529 & 0.000 & < 0.05 &  &         & 0.885 & 0.800 & < 0.05 \\ 
                                    & \textsc{OpenAI Classifier} &  &          & 0.631 & 0.120 & < 0.05 &  &         & 0.545 & 0.087 & < 0.05 &  &         & 0.503 & 0.010 & < 0.05 \\ \midrule
\multicolumn{17}{l}{\textbf{\textit{Watermarking}}}     \\
\multirow{6}{*}{\textbf{}} & \update{\rdfw} & & 28.3 & 0.605 & 0.091 & < 0.05 & & \textbf{32.1} & 0.486 & 0.024 & < 0.05 & & \textbf{21.0} & 0.497 & 0.020 & < 0.05 \\
                                    & \update{\rdfw~(w/ high entropy)} & & 16.7 & 0.749 & 0.402 & < 0.05 & & 14.3 & 0.828 & 0.512 & < 0.05 & & \textbf{21.0} & 0.513 & 0.040 & < 0.05 \\ \cmidrule(lr){4-17}
                                    & \wllm~($\Delta$\textsc{pass@1}~$\sim-10\%$)$^{\star}$          &  & 25.9 & 0.887 & 0.604 & < 0.05 &  & 25.5 & 0.833 & \textbf{0.518} & < 0.05 &  & 12.0 & 0.939 & 0.840 & < 0.05 \\
                                    & \myalgo~($\Delta$\textsc{pass@1}~$\sim-10\%$)$^{\star}$     &  & 26.2 & \textbf{0.943} & \textbf{0.817} & < 0.05 &  & 27.6 & \textbf{0.862} & 0.457 & < 0.05 &  & 13.0 & \textbf{0.980} & \textbf{0.920} & < 0.05  \\ \cmidrule(lr){4-17}
                                    & \wllm~(AUROC$\geq0.9$)$^{\dagger}$          &  & 25.9 & 0.887 & 0.604 & < 0.05 &  & 9.5 & 0.947 & 0.872 & < 0.05 &  & 12.0 & 0.939 & 0.840 & < 0.05 \\
                                    & \myalgo~(AUROC$\geq0.9$)$^{\dagger}$     &  & \textbf{29.0} & 0.904 & 0.707 & < 0.05 &  & 22.6 & 0.969 & 0.878 & < 0.05 &  & 13.0 & \textbf{0.980} & \textbf{0.920} & < 0.05 \\ \bottomrule
\end{tabular}%

}
\caption{\cam{\textbf{Main results} of code generation performance and detection ability on HumanEvalPack~\citep{muennighoff2024octopack} and ClassEval~\citep{du2023classeval}. Since calibration on watermarking strength leads to trade-offs between code generation quality and detection ability, we present two results for \update{$\wllm$ and $\myalgo$}. $^{\star}$ for the best detection score (i.e., AUROC and TPR) while allowing a code generation quality decrease of $\sim$10\% compared to Non-watermarked, and $^{\dagger}$ for the best code generation quality (\textsc{pass@1}) among AUROC \update{$\geq$} 0.9. We add $\rdfw$ and a Non-watermarked baseline with a high entropy setting (i.e., temperature=1.0 and top-p=1.0).
}
}
\label{tab:table_main_2_w_classeval}
\end{table*}


%% file: resources_acl/fig_pareto_frontier_main.tex
\begin{figure*}[hbt!] 
\begin{center}
\includegraphics[width=\textwidth]{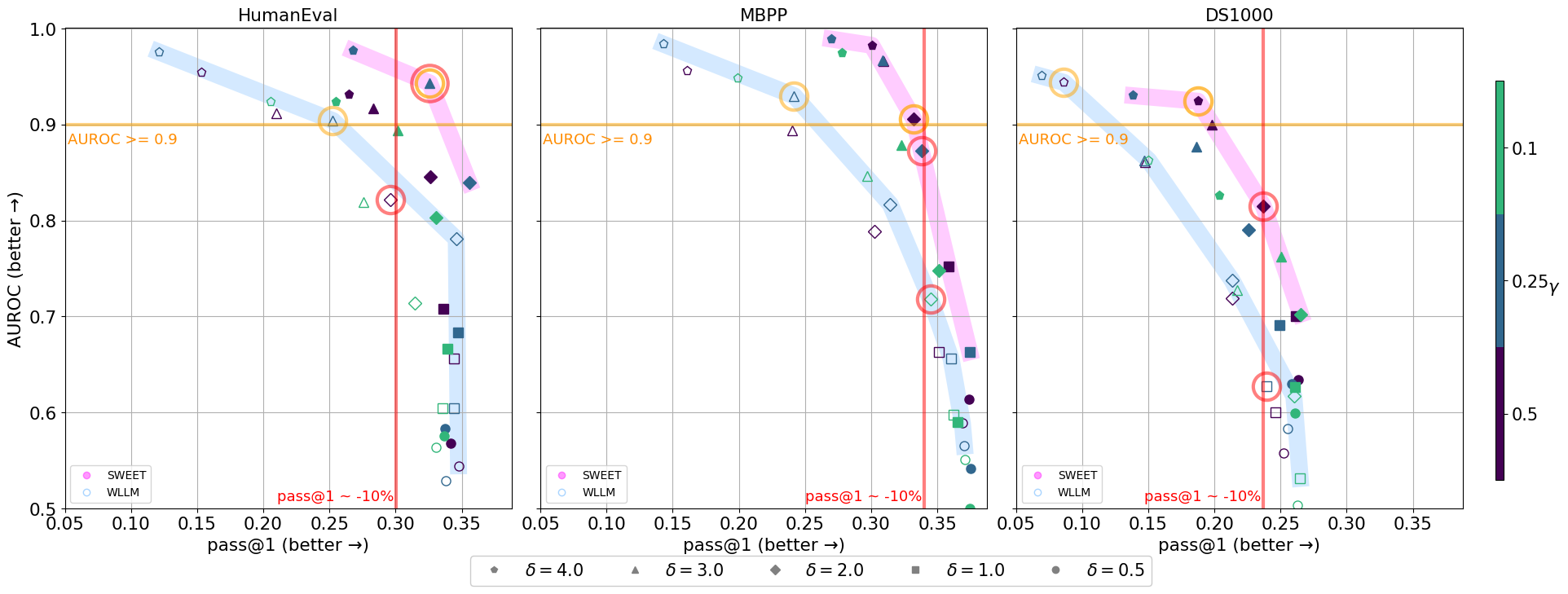}  
\end{center}
\caption{The tradeoff between AUROC and pass@1 of detecting real and generated samples of HumanEval, MBPP\update{, and DS-1000} datasets. The pink line represents a Pareto frontier of $\myalgo$, while the blue line represents that of $\wllm$. $\myalgo$ shows consistent dominance.
The \textcolor{red}{red}/\textcolor{orange}{orange} line and circles are the points used in Table~\ref{tab:table_main}. \update{The entropy threshold for $\myalgo$ is 1.2 here, and Pareto frontier figures for all threshold values are in Figure~\ref{fig:pareto_frontier_appendix}.}}
\label{fig:pareto_frontier_main}
\end{figure*}

%% file: sections/06_analysis.tex
\section{Analysis}\label{sec:analysis}

\input{resources_acl/fig_entropy_empact}

\subsection{Impact of Entropy Thresholds}
\label{ssec:threshold}

Figure \ref{fig:entropy} presents how code generation performance and detecting ability trade-off when calibrating the entropy threshold in our method. \update{$\wllm$ is when the entropy threshold is not applied (i.e., entropy threshold=0).} As the entropy threshold increases, the ratio of watermarked tokens decreases, so the code generation performance converges to a non-watermarked base model. This indicates that our method always lies between the $\wllm$ and a non-watermarked base model in terms of code generation performance. On the other hand, the detection ability, as the entropy threshold increases, reaches a local maximum but eventually declines. While our method with a moderate threshold effectively restricts generating the red list tokens compared to the $\wllm$, detection ability eventually decreases if the threshold is so high that few tokens are watermarked.
\cam{We further investigate how to effectively calibrate the entropy threshold value in Appendix~\ref{appendix:threshold_calibration}.}


\subsection{Detection Ability without Prompts
}\label{ssec:general_prompt}

As entropy information is required in the detection phase, approximating entropy values for each generation time step $t$ is essential in our method. In the main experiments, we prepend the prompt used in the generation phase (e.g., the question of Fig. \ref{fig:real_examples}) before the target code to reproduce the same entropy. However, we hardly know the prompt used for a given target code in the real world. Thus, instead of using the \textit{gold} prompt, we attach a common and general prompt for code generation to approximate the entropy information. We use five general prompts as below, and their z-scores are averaged for use in detection. \\

\input{resources_acl/resources_appendix/general_prompts}

Figure~\ref{fig:pareto_frontier_general_prefix_humaneval} demonstrates how the detection ability varies when using general prompts in the HumanEval dataset. $\myalgo$ with general prompts shows lower AUROC values than the original $\myalgo$, indicating inaccurately approximated entropy information impairs detection ability. Nevertheless, it still outperforms the $\wllm$ baseline regarding detection ability, drawing a Pareto frontier ahead of $\wllm$ \update{in all entropy threshold values}.

\subsection{\update{Use of Surrogate Model}}
\label{ssec:surrogate}
\update{
When detecting watermarks in a text, utilizing a smaller LM as a surrogate could be more computationally efficient and cost-effective~\citep{wang2023towards}.
We investigate the impact of employing this surrogate model during the detection phase. Specifically, we generate watermarked code using the original model (LLaMA2-13B) and detect watermarks using a smaller model (LLaMA2-7B).
}

\update{In the results of Figure~\ref{fig:pareto_frontier_llama_humaneval}, the detection performance declines are insignificant, and our approach utilizing the surrogate model continues to surpass the baseline. Such performance preservation may be due to that LLaMA2 7B and 13B are trained on the identical training corpus \citep{touvron2023llama2}. Further analysis for computational cost can be found in Appendix~\ref{appendix:computation_cost}.}

\subsection{\update{Robustness to Paraphrasing Attacks}}
\label{ssec:robustness}
\input{resources_acl/fig_paraphrasing}
Even with the text watermarked, a malicious user might attempt to remove watermarks in the text by paraphrasing \citep{krishna2023paraphrasing,sadasivan2023can}. Paraphrasing the code text is more restrictive than dealing with plain text because it must avoid triggering any code malfunctions. We assess the robustness of watermarking methods against paraphrasing by employing \cam{two types of attacks - changing the names of variables and utilizing a commercial code refactoring service.\footnote{https://codepal.ai/code-refactor}} Specifically, \cam{for each watermark method, we choose 273 source codes from the MBPP task, for which all three methods succeed in generating with no syntax error. In the code renaming attack,} we select variables in the watermarked code and rename them with randomly generated strings of varying lengths, ranging from 2 to 5 characters. \cam{We use five random seeds for renaming.}

\update{Figure~\ref{fig:paraphrasing} presents the results of the detection performance on the \cam{paraphrased code.} All watermarking methods show the decline of AUROC scores when \cam{the extent of paraphrasing increases}, while our approaches continue to show better performances than baselines.
However, our approaches also show that the AUROC scores drop to about 0.8 when all variables are renamed. We found that this is because variable names comprise a large proportion of high entropy tokens in the code text (See Appendix~\ref{appendix:lexical} for details).
}

%% file: resources_acl/fig_entropy_empact.tex
\begin{figure}[t!]
\begin{center}
\includegraphics[width=\linewidth]{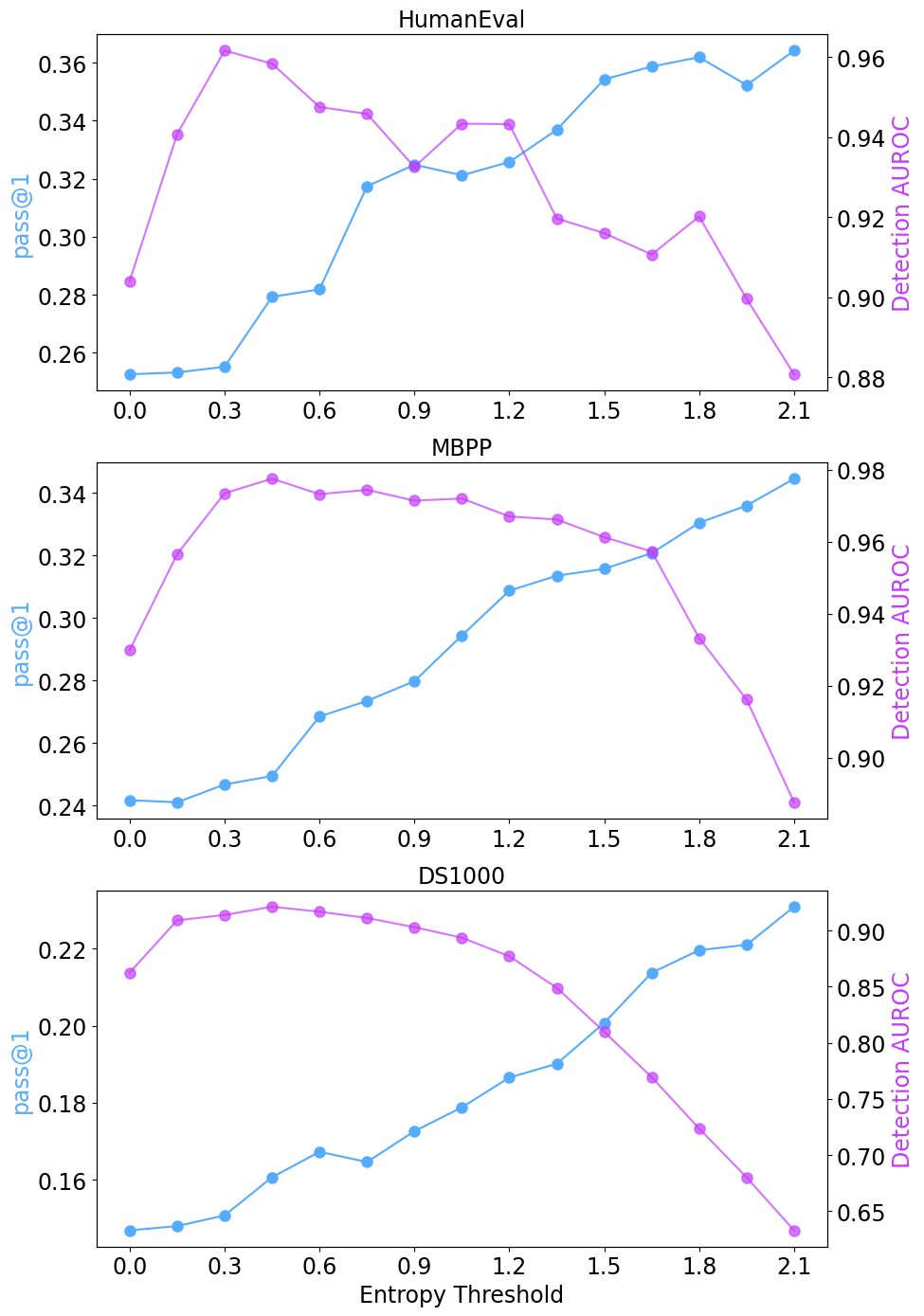}  
\end{center}
\caption{
Plots of code quality pass@1 and detection AUROC when calibrating the entropy threshold of our methods, $\myalgo$, on the three code benchmarks. We set $\gamma=0.25$ and $\delta=3.0$. While code generation performance increases with a higher entropy threshold, detection AUROC scores make an up-and-down curve.
}\label{fig:entropy}
\end{figure}

%% file: resources_acl/resources_appendix/general_prompts.tex
\noindent\begin{minipage}{\linewidth}
\begin{lstlisting}
def solution(*args):
    """
    Generate a solution
    """
\end{lstlisting}
\end{minipage}
\noindent\begin{minipage}{\linewidth}
\begin{lstlisting}
<filename>solutions/solution_1.py
# Here is the correct implementation of the code exercise
def solution(*args):
\end{lstlisting}
\end{minipage}
\noindent\begin{minipage}{\linewidth}
\begin{lstlisting}
def function(*args, **kargs):
    """
    Generate a code given the condition
    """
\end{lstlisting}
\end{minipage}
\noindent\begin{minipage}{\linewidth}
\begin{lstlisting}
from typing import List

def my_solution(*args, **kargs):
    """
    Generate a solution
    """
\end{lstlisting}
\end{minipage}
\noindent\begin{minipage}{\linewidth}
\begin{lstlisting}
def foo(*args):
    """
    Solution that solves a problem
    """
\end{lstlisting}
\end{minipage}

%% file: resources_acl/fig_paraphrasing.tex
\begin{figure}[t!]
\begin{center}
\includegraphics[width=\linewidth]{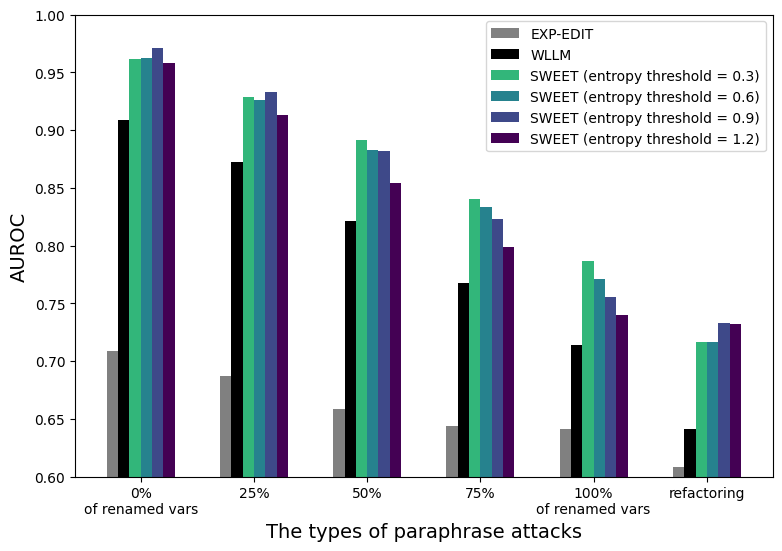}  
\end{center}
\caption{
Watermark detection performance on renamed variables in the code.
We set $\gamma=0.25$ and $\delta=3.0$ for $\wllm$ and $\myalgo$. For $\rdfw$, we search the hyperparameter for the block size in [20,30,40] with a high entropy setting.
}\label{fig:paraphrasing}
\end{figure}

%% file: sections/07_conclusion.tex
\section{Conclusion}\label{sec:conclusion}
We identified and emphasized the need for Code LLM watermarking, and formalized it for the first time.
Despite the rapid advance of coding capability of LLMs, the necessary measures to encourage the safe usage of \update{code generation} models have not been implemented yet.
Our experiments showed that existing watermarking and detection techniques failed to properly operate under the code generation setting.
The failure occurred in two modes: either 1) the code does not watermark properly (hence, cannot be detected), or 2) the watermarked code failed to properly execute (degradation of quality).
Our proposed method SWEET, on the other hand, improved both of these failure modes to a certain extent by introducing selective entropy thresholding which filters tokens that are least relevant to execution quality.
\update{In code generation tasks, our method performs better than baselines, including post-hoc detection methods, while achieving less code quality degradation.
Moreover, comprehensive analysis demonstrates that our method still works well in real-world settings, specifically when the prompts are not given, utilizing even a smaller surrogate model, or under paraphrasing attacks.}

%% file: sections/08_limitations.tex
\section*{Limitations}\label{sec:limitation}

We identify the limitations of this work and suggest ways to mitigate them. First, two issues are shared by the status quo of this field as follows. (1) Robustness against paraphrasing attacks: As users can tailor LLM's code to their specific needs, it is crucial to be robust against paraphrasing attacks. We addressed this issue in Section~\ref{ssec:robustness} and left further robustness enhancement for future work. \cam{(2) Possibilities of watermark forgery: An attacker may pry out the watermarking rules, and $\mathcal{O}(|\mathcal{V}|)^2$ runs in the brute-force mechanism enable it. Against the attack, one can apply techniques enhancing the watermarking model’s security, such as dividing the green/red list depending on prior $h>1$ tokens, as mentioned in the \wllm~paper, or applying methods like SelfHash~\cite{kirchenbauer2023reliability}.}

For our work, two \cam{additional} issues exist as follows. \cam{(1) Entropy threshold calibration: We demonstrate that our method outperforms the baselines in the broad entropy threshold range (see Sec~\ref{ssec:threshold}) and investigate how to calibrate the entropy threshold effectively (see Appendix~\ref{appendix:threshold_calibration}). However, we still need entropy threshold tuning to obtain the best performance, which costs a computation.} (2) Need for the source LLM during detection: \myalgo~works in a white-box setting. Although it has been shown that employing even a smaller surrogate LM can still maintain the detection performances to some degree (see Sec~\ref{ssec:surrogate}), this can be a computational burden for some users who want to apply our work.

%% file: sections/ethical.tex
\section*{\update{Ethical Statement}}\label{sec:ethical}
\update{Although watermarking methods are designed to address all potential misuse of LLMs by detecting machine-generated texts, they can simultaneously pose a new risk. For example, if a watermarking mechanism for a specific LLM is leaked to the public, a malicious user aware of this mechanism could abuse the watermarks to create unethical texts embedded with the model's watermarks. To prevent such scenarios, we recommend that all users exercise caution to avoid exposing the detailed mechanism, such as the key value for the hash function used to divide green and red lists in our method.}

%% file: sections/ack.tex
\section*{Acknowledgements}\label{sec:ack}
\cam{We would like to thank the anonymous reviewers for their valuable feedback.
This work was supported by 
SNU-NAVER Hyperscale AI Center, 
Institute of Information \& Communications Technology Planning \& Evaluation (IITP) grant funded by the Korea government (MSIT) (No.~RS-2019-II191082, SW StarLab), 
the National Research Foundation of Korea (NRF) grant funded by the Korea government (MSIT) (No.~2023R1A2C2005573), 
and the Basic Science Research Program through the National Research Foundation of Korea(NRF) funded by the Ministry of Education(RS-2023-00274280). 
}

%% file: sections/appendix.tex
\clearpage
\appendix

\section{Preliminaries for~\wllm}\label{appendix:preliminaries}

In this section, we provide brief preliminaries for~\citet{Kirchenbauer2023watermark}.
For a given language model $f_{\text{LM}}$ with vocabulary $\vocab$,
the likelihood probability of a token $y_{t}$ is calculated as follow:
\begin{equation}\label{eq:logit}
\vl_{t}=f_{\text{LM}}(\bm{x}, \bm{y}_{[:t]}),
\end{equation}
\begin{equation}\label{eq:prob}
p_{t,i}=\frac{e^{\vl_{t}^{i}}}{\sum_{i=1}^{|\vocab|}e^{\vl_{t}^{i}}},
\end{equation}
where $\bm{x}=\{x_0,\dots,x_{M-1}\}$ and $\bm{y}_{[:t]}=\{y_1,\dots,y_{t-1}\}$ are a $M$-length tokenized prompt and the generated token sequence, respectively, and $\vl_t\in\mathbb{R}^{|\vocab|}$ is the logit vector.

\paragraph{Watermarking in LM-generated Text.}

In the watermarking \citep{Kirchenbauer2023watermark}, the entire tokens in $\vocab$ at each time-step are randomly binned into the green $\mathcal{G}_t$ and red groups $\mathcal{R}_t$ in proportions of $\gamma$ and $1-\gamma$ $(\gamma \in (0,1))$, respectively. The method increases the logits of green group tokens by adding a fixed scalar $\delta$, promoting them to be sampled at each position. Thus, watermarked LM-generated text is more likely than $\gamma$ to contain the green group tokens.
On the other hand, since humans have no knowledge of the hidden green-red rule, the proportion of green group tokens in human-written text is expected to be close to $\gamma$.

The watermarked text is detected through a one-sided $z$-test by testing the null hypothesis where the text is not watermarked.
The $z$-score is calculated using the number of recognized green tokens in the text.
Then, the testing text is considered as watermarked if the $z$-score is greater than $z_{\text{threshold}}$.
Note that the detection algorithm with the higher $z_{\text{threshold}}$ can result the lower false positive rate (FPR) and reduce Type I errors.

\textbf{Spike Entropy}
\citet{Kirchenbauer2023watermark} used \textit{spike entropy} for measuring how spread out a distribution is. Given a token probability vector $p$ and a scalar $m$, spike entropy of $p$ with modulus $m$ is defined as:

\begin{equation}\label{eq:spike_entropy}
S(p,m)=\sum_{}^{}\frac{p_k}{1+mp_k}.
\end{equation}

\input{resources_acl/resources_appendix/alg_generation}
\input{resources_acl/resources_appendix/alg_detection}

\section{Watermark Embedding/Detecting Algorithm of~\myalgo}\label{appendix:algorithm}
Algorithms~\ref{alg:alg1} and ~\ref{alg:alg2} show the detailed steps of generating a watermark and later detecting it using our selective entropy thresholding method (\myalgo).

Instead of the spike entropy used in $\wllm$, we use the classical Shannon entropy.
Given a token probability distibution vector $p$, the entropy of $p$ is computed by
\begin{equation}\label{eq:entropy}
H_t=-\sum_{}^{}p_{k}\log p_{k}.
\end{equation}

\section{Proof of Theorem~\ref{theorem:1}}\label{appendix:theorem}

We begin with a lemma from~\citet{Kirchenbauer2023watermark}, which predicts the probability of a green list token sampled from a language model employing the watermarking.

In our proof, we predict the lower bounds of $z$-score when detecting watermarks via $\wllm$ or $\myalgo$ methods and compare the z-score lower bounds.

\begin{lemma}\label{lemma:1}
Suppose $p \in (0,1)^{|\vocab|}$ is a raw probability vector generated from a language model where $|\vocab|$ is the vocabulary size.
Before sampling $p$, watermarks are embedded by dividing randomly a green list of size $\gamma |\vocab|$ and a red list of size $(1-\gamma)|\vocab|$ for some value $\gamma \in (0,1)$. It then promotes the logits of tokens in the green list by $\delta$. When sampling a token index $k$ from this watermarked distribution, the probability that the token is sampled from the green list (considering the randomness of green list) is at least
\begin{equation}\label{eq:green_lower_bound}
\mathbb{P}[k \in G] \ge \frac{\gamma e^\delta}{1+(e^\delta-1)\gamma}S(p, \frac{(1-\gamma)(e^\delta-1)}{1+(e^\delta-1)\gamma}).\nonumber
\end{equation}
\end{lemma}

Let's begin the proof.

\begin{proof}

In $\wllm$, we consider all tokens in $\bm{y}=\{y_0,\dots,y_{N-1}\}$ for detection. We can get a lower bound of the number of green list tokens in $\bm{y}$ by summing the result of Lemma~\ref{lemma:1} over the tokens $y_t$. The expectation of the number of green list tokens, $N_g$, in $\bm{y}$ is at least
\begin{equation}\label{eq:wllm_green_number}
\mathbb{E}[N_G] \ge \alpha\gamma N \overline{S}.
\end{equation}
where $\alpha=\frac{e^\delta}{1+(e^\delta-1)\gamma}$, and $\overline{S}=\sum_{t=1}^N S_t/N$.

We can get the lower bound of the $z$-score by applying the $z$-score definition in Eq.~\ref{eq:z-stat}:
\begin{equation}\label{eq:wllm_z_stat}
z \ge \gamma\sqrt{N}\frac{\alpha\overline{S}-1}{\sqrt{\gamma(1-\gamma)}}.
\end{equation}

If the entropy threshold is applied, we consider only tokens with entropy values higher than the threshold to be tested. Let $N^h$ be the number of tokens that have higher entropy values.
Following Eq.~\ref{eq:wllm_green_number} and Eq.~\ref{eq:wllm_z_stat} again with $N^h$, we can get the lower bound of the $z$-score of $\myalgo$:


\begin{equation}\label{eq:sweet_z_stat}
z \ge \gamma\sqrt{N^h}\frac{\alpha\overline{S^h}-1}{\sqrt{\gamma(1-\gamma)}},\nonumber
\end{equation}
where $\overline{S^h}=\sum_{t=1}^N S_t \times \mathbb{1}(S_t \ge \threshold) / N^h$.

$\overline{S_h} \ge \overline{S}$ is ensured as we ignore all tokens with lower entropy than the threshold. By comparing Eq.~\ref{eq:wllm_z_stat} and Eq.~\ref{eq:sweet_z_stat},

\begin{align}
\gamma\sqrt{N^h}\frac{\alpha\overline{S^h}-1}{\sqrt{\gamma(1-\gamma)}} &\ge \gamma\sqrt{N}\frac{\alpha\overline{S}-1}{\sqrt{\gamma(1-\gamma)}},\nonumber\\
\sqrt{\frac{N-N^l}{N}} &\ge \frac{\alpha\overline{S}-1}{\alpha\overline{S^h}-1},\nonumber\\
\frac{N^l}{N} &\le 1-(\frac{\alpha\overline{S}-1}{\alpha\overline{S^h}-1})^2,\nonumber
\end{align}
where $N^l=N-N^h$.
\end{proof}

\input{resources_acl/resources_appendix/tab_llama_humaneval}

\section{Implementation Details}\label{appendix:Implementation Details}

\update{We have used three datasets for our testbeds: HumanEval, MBPP, and DS-1000. They have 164, 500, and 1000 Python code problems, respectively.}
For our base models, StarCoder and LLaMA\update{2}, we use top-$p$ \citep{Holtzman2020topp} sampling with $p=0.95$ for both models, and temperature 0.2 and 0.1, respectively. When generating output for each code problems, we use zero-shot setting in HumanEval \update{and DS-1000} but 3-shot in MBPP. Prompts used in MBPP are similar to the prompt in~\citet{austin2021MBPP}. For calculating pass@1 scores, we set $n=40$ for HumanEval and DS-1000, and $n=20$ for MBPP. \cam{We use a single NVIDIA RTX A6000 GPU to generate or detect each code completion with StarCoder or LLaMA2. It takes less than two GPU hours for generation and less than 1 GPU hour for detection.}

\subsection{DetectGPT}
We used two masking models for DetectGPT. When T5-3B is used for DetectGPT, we search hyperparameters for the length of the spans in [1,2,5,10] words, and for the proportion of masks in [5,10,15,20]\% of the text. When utilizing SantaCoder, we simulate the single-line fill-in-the-middle task scenario by masking only one line of code per perturbation, which is a task that SantaCoder is trained to perform well. \citep{fried2023incoder,bavarian2022efficient}. We search hyperparameters for the number line to be rephrased in [1,2,3,4]. We make 100 perturbations following the original paper.

\subsection{\wllm~and~\myalgo}
Depending on the strength of watermark, trade-off between code functionality and watermarking detectability exists. We search hyperparameters for the ratio of the green list $\gamma$ in [0.1,0.25,0.5], and for the green token promotion value $\delta$ in [0.5,1.0,2.0,3.0,4.0]. \update{For the entropy threshold values used in $\myalgo$, we search thresholds in [0.3,0.6,0.9,1.2].}

\update{\subsection{\rdfw}
In most tasks we have conducted experiments, the length of the generated code hardly exceed 100 tokens. Therefore, considering that length of the watermark key sequence significantly affected the detection speed, we search hyperparameters for the length of the key sequence only in [100, 500]. The block size was set equal to the length of the model output, and the resample size $T=500$ for all instances. To generate $n$ outputs to calculate pass@k, we shift the watermark key sequence randomly $n$ times. Finally, we set edit distance hyperparameter $\gamma=0.0$ for $\rdfw$ as used in their paper.}

\section{Experimental Details and Results on HumanEvalPack and ClassEval}\label{appendix:humanevalpack_classeval}
\cam{We choose two additional benchmarks to present how well our approach functions in broader software development contexts.}

\subsection{HumanEvalPack}
\cam{HumanEvalPack~\citep{muennighoff2024octopack} is an extension of HumanEval to cover 6 languages (Python, C++, Java, Javascript, Go, and Rust), and we choose C++ and Java as our testbeds. Basically, all hyperparameter settings are equal to Python benchmarks except $n=5$. For \wllm~and \myalgo, we narrow the search space for $\delta$ into [1.0,2.0,3.0,4.0]. For \rdfw, we fix the length of the key sequence to 100.}

\subsection{ClassEval}
\cam{ClassEval~\citep{du2023classeval} differs from the aforementioned datasets in terms of code generation scopes as it requires language models to generate class-level code passages rather than just a single function. It consists of 100 class-level code generation test examples in which a model has to generate a whole Python class code given a skeleton code of the class. Following the ClassEval paper, we use StarCoder-Instruct~\citep{GeorgiaTechResearchInstitute2023starcoderinstruct} as the base model and the same instruction-following template of the prompt used in the paper. As the entropy distribution from StarCoder-Instruct skews to lower values than that of the StarCoder model, we search hyperparameters for threshold in [0.01,0.03,0.05,0.1,0.2] and $\delta$ in [2,3,4,5,10,15,20]. We generate $n=5$ outputs to calculate pass@5. For \rdfw, due to the high computational cost, we make 50 perturbations instead of 100.}

\subsection{Results}
\cam{As presented in Table~\ref{tab:table_main_2_w_classeval}, \myalgo~outperforms all baselines in detecting machine-generated code in C++ and Java environments. Also, it still preserves code functionality much more than \wllm~while achieving better detection performance. We observe that the Pareto Frontier lines of~\myalgo~are ahead of those of~\wllm, as in the Python environment (see Figure~\ref{fig:pareto_frontier_cpp_java}). It is worth noting that the C++ and Java examples in HumenEvalPack comprise longer code than the Python examples we used in the paper: an average of 100 tokens for C++, 97 tokens for Java, and 57 tokens for Python. Therefore, these results demonstrate that the efficacy of our methodology is not limited to the type of programming languages or the length of the code.}

\cam{In the class-level code generation task, we could still observe our approach showing the highest AUROC score than the baselines even in the class-level code generation task where the LLM should generate longer and more complex code (the average token length of ClassEval solutions is 352). Specifically, \myalgo~achieves an AUROC of 0.980 with a 7.1\% degradation of code functionality. Interestingly, as the text becomes longer, post-hoc methods' performance increases. On the other hand, \rdfw~has shown lower detection performance than when evaluated in StarCoder, even in the high-entropy setting, due to the extremely spiky entropy distribution of StarCoder-Instruct.}
\cam{In addition, \rdfw~increases code functionality compared to non-watermarked baselines, even though it is a watermarking method that does not distort the original token distribution. We suppose these results are attributed to the specific watermark key sequence, which is randomly generated.}

\section{Further Pareto Frontier Results on StarCoder/LLaMA2}\label{appendix:pareto}
\input{resources_acl/resources_appendix/fig_pareto_frontier_appendix}
\input{resources_acl/resources_appendix/fig_pareto_frontier_cpp_java}
\input{resources_acl/resources_appendix/fig_pareto_frontier_general_prompt}
\input{resources_acl/resources_appendix/fig_pareto_frontier_temp0.8}
\input{resources_acl/resources_appendix/fig_pareto_frontier_llama2_and_surrogate}

\textbf{HumanEval pass@100.} Figure \ref{fig:pareto_frontier_humaneval_0.8_auroc} shows a tradeoff between pass@100 score and AUROC at HumanEval task in temperature 0.8. We generated 200 samples in HumanEval to calculate pass@100. The tendency of the Pareto Frontier are the same, $\myalgo$ is consistently placed in the front. While pass@100 score is much higher than the pass@1 score at temperature=0.2, we see the range of AUROC remains similar. This indicates temperature does not affect the detection strength of each samples heavily.


\textbf{LLaMA\update{2}.} Furthermore, Table~\ref{tab:llama_humaneval} shows the results on HumanEval when using LLaMA\update{2 13B} (a general-purpose LLM), as the backbone for code generation. We can observe similar trends as demonstrated in Figure~\ref{fig:pareto_frontier_llama_humaneval}.
$\myalgo$ in LLaMA\update{2} achieves a higher AUROC than all other baselines while preserving code quality more than $\wllm$.
Consequently, we observe that $\myalgo$ also applies to general-purpose LLM, which is not code-specific.

\section{\update{Detectability with Varying Code Lengths}}
\label{appedix:detectability@T}
\input{resources_acl/resources_appendix/fig_detectability_appendix}
\update{We experiment the detection performance across different code lengths. Based on the detectability@T metric proposed in \citet{kirchenbauer2023reliability}, we evaluate the detection performance within the first $T$ tokens of the machine-generated and human-written code sequences and calculate AUROC scores.}

\update{As presented in Figure~\ref{fig:detectability_appendix}, $\myalgo$ demonstrates superior detection performance even in the short code texts. This is particularly important feature in code generation tasks comprised of relatively shorter texts than plain text generation. Moreover, in HumanEval and MBPP, we can observe that the AUROC of $\myalgo$ reaches 1.0 with the text length exceeding 70, while none of the baselines could achieve it.}

\section{Entropy Threshold Calibration}\label{appendix:threshold_calibration}

\input{resources_acl/resources_appendix/fig_entropy_calibration}

\cam{This section proposes a method to calibrate the best entropy threshold effectively.}
\cam{To detect machine-generated texts, the z-score of them should be high. Let’s say there is a watermarked text sequence with an entropy threshold $\tau$, and the length is $N$. The z-score is calculated as in Equation~\ref{eq:z-stat}. Thus $z \propto \sqrt{\frac{N^h}{N}}(\frac{N^h_G}{N^h}-\gamma)$, assuming a fixed $N$. If we can find a relationship between $\tau$ and $\frac{N^h}{N}$, and $\tau$ and $\frac{N^h_G}{N^h}$, we could choose $\tau$ that maximizes the z-score in a fixed $N$. We denote $z'$ as a pseudo-metric for estimating z-score:}

\begin{equation}\label{eq:z'}
    z’=E[\frac{N^h}{N}] (E[\frac{N^h_G}{N^h}]-\gamma)
\end{equation}

\cam{With logits generated by an LLM, we can calculate entropy $H$ and the probability of sampling a green token after adding $\delta$ to green tokens' logits. We call the expectation of it over the randomness of green/red list partitioning as $P_G$. We model the distribution of logits that LLM generates as a probability distribution function $P(H, P_G; \gamma, \delta)$. We approximate the tokens in a text sequence as i.i.d, then we can write as follows:}

\begin{equation}
    E[\frac{N^h}{N}] = P(H > \tau, P_G)\nonumber
\end{equation}
\begin{equation}
    E[\frac{N^h_G}{N^h}]=E[P_G|H>\tau]\nonumber
\end{equation}

\cam{
To estimate $P(H, P_G; \gamma, \delta)$ in the Python language domain of StarCoder, we use a code corpus, CodeSearchNet~\citep{husain2019codesearchnet}. Specifically, we feed the Python corpus of CodeSearchNet to our model and obtain all logits for each time step and calculate $H$ and $P_G$. For $P_G$, we averaged the probability of sampling a green token from a watermarked distribution with 500 random green/red list partitions. We regard the pair ($H$, $P_G$) per one logits as an unnormalized joint discrete distribution.}

\cam{The results are presented in Figure~\ref{fig:threshold_calibration}. The $z’$ is the highest when the entropy threshold is in $[0.820, 0.871]$. It aligns with the results in Figure~\ref{fig:entropy}, where the optimal threshold value lies around 0.3$\sim$0.9. Therefore, the threshold value found here is a good starting point for searching for the optimal threshold value. The computational cost is only one forward pass across the corpus we used. The result indicates that we can use the information in the code corpus to calibrate an entropy threshold effectively.}

\section{Analysis of Computation Cost}\label{appendix:computation_cost}
\update{It is practically important to detect machine-generated text without a huge computational overload. We here analyze computation costs for each baseline and our method.}

WLLM does not require any additional computation as it only needs a random number generator and a seed number to put. On the other hand, all zero-shot post-hoc detection methods excluding DetectGPT need at least one forward pass of that LLM. DetectGPT needs to run forward passes as much as the number of perturbations for increased accuracy (the original paper generated 100 perturbed samples, so we did the same). Our method needs one time forward pass to calculate the entropy, which is the same with zero-shot post-hoc detection methods except for DetectGPT. \update{However, we demonstrated that our method outperforms baselines even when utilizing a smaller surrogate model (Sec~\ref{ssec:surrogate}), indicating the capability of computationally more efficient employment.}
\cam{On the other hand, while $\rdfw$ does not need LLM for detecting watermarks, it requires measuring the Levenshtein distance to compute the test statistic. Specifically, it demands an extensive calculation of $O(mnk^2)$, where $m$ be the length of the target text, $n$ be the length of the watermark key sequence, and $k$ be the block size.} Moreover, $T=500$ times of test statistic is also necessary for reporting the p-value. Although these computations do not require LLM and can be implemented in parallel, one can consider the computation cost of $\rdfw$ as high.

\section{Analysis of Lexical Type Distributions}\label{appendix:lexical}

\input{resources_acl/resources_appendix/fig_lexical_distribution_above_entropy}
\input{resources_acl/resources_appendix/fig_lexical_distribution_below_entropy}

Watermarking a text without degrading its quality is possible when many candidates are alternatively available. In code generation, it is challenging to achieve this, so $\myalgo$ selectively apply watermarking only on high entropy, i.e., when there are many candidates. Using Python built-in tokenize module\footnote{\url{https://docs.python.org/3/library/tokenize.html}}, we here tokenize outputs of our $\myalgo$ method and analyze the distributions of lexical types both above and below the entropy threshold.

\subsection{List of Lexical Types}

Below is the list of lexical types we use for analysis and corresponding examples. All list of types the tokenize module actually emits can be found in \url{https://docs.python.org/3/library/token.html}. We merged and split the original types.

\begin{itemize}

    \item NAME : identifier names, function names, etc.
    \item OP : operators, such as \{, [\, (\, +, =, etc.
    \item INDENT : we merge NEWLINE, DEDENT, INDENT, NEWLINE, and NL.
    \item RESERVED : split from NAME. In Python docs, they are officially named \textit{keywords}.
    \item BUILT-IN : split from NAME. Please refer to Python docs\footnote{\url{https://docs.python.org/3/library/functions.html\#built-in-functions}}.
    \item NUMBER
    \item STRING
    \item COMMENT
    \item FUNCNAME : split from NAME. We manually build a list of function name almost being used only for function. For examples, append(), join(), split() functions are included.

\end{itemize}

\subsection{Lexical Types Distributions Above Threshold}

Figure~\ref{fig:lex_dis_above_ent} shows lexical types distributions of output tokens above the entropy threshold (i.e., watermarked tokens) across seven thresholds. As the entropy threshold rises, the proportion of NAME type tokens increases by the most (26\%p to 63\%p). Intuitively, this can be easily understood, considering there would be many alternative candidates for defining identifier names. Unfortunately, this would lead to vulnerability to an adversarial attack on watermarking, such as changing variable names. Following the NAME type, the ratio of the RESERVED type also increases slightly (12\%p to 20\%p), meaning that the model has multiple choices of logical flow in code generation, considering RESERVED tokens usually decide code execution flow.

\subsection{Lexical Types Distributions Below Threshold}

Figure~\ref{fig:lex_dis_below_ent} shows lexical types distributions of output tokens below the entropy threshold. In contrast to the distributions above the threshold, NAME and RESERVED types do not increase as the threshold rises. Meanwhile, the proportion of INDENT types slightly increases (18\%p to 22\%p), indicating that the model has more confidence in the rules, such as indentation.

\section{Further Analysis of Breakdown of Post-hoc methods}
\label{appendix:breakdown}

The performance of post-hoc detection methods in the machine-generated code detection task is surprisingly low compared to their performance in the plain text domain. In both HumanEval and MBPP, none of the post-hoc baselines have an AUROC score exceeding 0.6, and the TPR is around 10\% or even lower. In this section, we analyze the failures of post-hoc detection baselines.

\textbf{Out-Of-Domain for classifiers}.
Methods leveraging trained classifiers, such as GPTZero and OpenAI Classifier, inherently suffer from out-of-domain (OOD) issues~\citep{Guo2023How,Yang2023DNA}. Since the machine-generated code detection problems are relatively under explored, we can conjecture that there are not enough examples of machine-generated code for training, especially even though we do not know of the dataset on which GPTZero was trained.

\textbf{Relatively Short Length of Code Blocks}.
DetectGPT presumes the length of the text being detected as near paragraph length. OpenAI Classifier released in 2023~\citep{OpenAI2023classifier} takes only text longer than 1,000 tokens. Even in the WLLM and their following paper~\citep{kirchenbauer2023reliability}, the length is one of the prime factors in detection and is used in a metric, detectability@T. Despite the importance of the length, in our experiments, the length of the generated code text is generally short. The token lengths generated by the model were are 59 and 49 tokens on average for HumanEval and MBPP, respectively. Unless embedding some signals in the text intentionally, like WLLM and ours, it seems that it is challenging for post-hoc methods to detect short text.

\textbf{Failures in DetectGPT}.
Specifically, in DetectGPT, we attribute the failure to detect machine-generated code to poor estimation of perturbation curvature. We hypothesize two reasons for this. Firstly, considering the nature of the code, it is challenging to rephrase a code while preserving its meaning or functionality. To minimize the degradation of perturbation, we use SantaCoder for the masking model and paraphrase only one line of code at a time. Yet, in most cases, the rephrased code is either identical to its original or broken in functionality. Secondly, LLMs have not achieved as satisfactory code generation performance as plain text generation. Hence, the base and masking models cannot draw meaningful curvature.

%% file: resources_acl/resources_appendix/alg_generation.tex
\begin{algorithm}[t]
\caption{Generation Algorithm of \myalgo}\label{alg:alg1}
\begin{algorithmic}[1]
\STATE{\bfseries Input:} tokenized prompt $\bm{x}=\{x_1,\dots,x_{M-1}\}$; entropy threshold $\threshold\in[0,\log|\mathcal{V}|]$, $\gamma\in(0,1)$, $\delta>0$;
\FOR{$t=0,1,2,\dots$}
\STATE Compute a logit vector $\vl_t$ by \eqref{eq:logit};
\STATE Compute a probability vector $\vp_t$ by \eqref{eq:prob};
\STATE Compute an entropy $H_t$ by \eqref{eq:entropy};
\IF{$H_t>\threshold$}
\STATE Compute a hash of token $y_{t-1}$, and use it as a seed for a random number 
generator;
\STATE Randomly divide $\vocab$ into $\mathcal{G}_t$ of size $\gamma|\mathcal{V}|$ and $\mathcal{R}_t$ of size $(1-\gamma)|\mathcal{V}|$;
\STATE Add $\delta$ to the logits of tokens in $\mathcal{G}_t$;
\ENDIF
\STATE Sample $y_t$;
\ENDFOR
\end{algorithmic}
\end{algorithm}

%% file: resources_acl/resources_appendix/alg_detection.tex
\begin{algorithm}[ht!]
\caption{Detection Algorithm of \myalgo}\label{alg:alg2}
\begin{algorithmic}[1]
\STATE{\bfseries Input:} tokenized prompt $\bm{x}$; token sequence to be tested $\bm{y}=\{y_0,\dots,y_{N-1}\}$; entropy threshold $\threshold\in[0,\log|\mathcal{V}|]$, $\gamma\in(0,1)$, $z_{\text{threshold}}>0$;
\STATE Set $N^h = 0$ and $N^h_G = 0$;
\FOR{$t=0,1,2,\dots N-1$}
\STATE Compute a logit vector $\vl_t$ by \eqref{eq:logit};
\STATE Compute a probability vector $\vp_t$ by \eqref{eq:prob};
\STATE Compute an entropy $H_t$ by \eqref{eq:entropy};
\IF{$H_t>\threshold$}
\STATE $N^h \gets N^h+1$;
\STATE Compute a hash of token $y_{t-1}$, and use it as a seed for a random number generator;
\STATE Recover $\mathcal{G}_t$ and $\mathcal{R}_t$;
\IF{$y_t\in\mathcal{G}_t$}
\STATE $N^h_G \gets N^h_G+1$;
\ENDIF
\ENDIF
\ENDFOR
\STATE Compute $z$-score by \eqref{eq:z-stat};
\IF{$z>z_{\text{threshold}}$}
\STATE \textbf{return} True; (i.e., $\bm{y}$ is watermarked)
\ELSE
\STATE \textbf{return} False;
\ENDIF
\end{algorithmic}
\end{algorithm}

%% file: resources_acl/resources_appendix/tab_llama_humaneval.tex
\begin{table}[t]
\centering
\tiny
\begin{tabular}{@{}lllll@{}}
\toprule
\multicolumn{1}{c}{\multirow{2}{*}{Method}} & \multicolumn{4}{c}{HumanEval} \\ \cmidrule(l){2-5} 
\multicolumn{1}{c}{} & pass@1 & AUROC & TPR & FPR \\ \midrule
Non-watermarked & \update{\textbf{17.3}} & - & - & - \\
\update{Non-watermarked (w/ high entropy)} & 6.8 & - & - & - \\ \midrule
\update{\rdfw} & \textbf{17.1} & 0.612 & 0.110 & <0.05 \\
\update{\rdfw~(w/ high entropy)} & 7.1 & 0.844 & 0.561 & <0.05 \\ \midrule
\wllm~($\Delta$\textsc{pass@1}~$\sim-10\%$)$^{\star}$ & 15.4 & 0.777 & 0.402 & <0.05 \\
\myalgo~($\Delta$\textsc{pass@1}~$\sim-10\%$)$^{\star}$ & 15.5 & \textbf{0.921} & \textbf{0.616} & <0.05 \\ \midrule
\wllm~(AUROC$\geq0.9$)$^{\dagger}$ & 9.2 & 0.908 & 0.720 & <0.05 \\
\myalgo~(AUROC$\geq0.9$)$^{\dagger}$ & 15.5 & 0.921 & 0.616 & <0.05 \\ \bottomrule
\end{tabular}
\caption{Results of code generation performance and detection ability in LLaMA\update{2 13B}. We calculate pass@1 metrics by generating $n=40$ examples. Hyperparameters for decoding strategy is top-p decoding with $p=0.95$ and temperature=\update{$0.1$}\update{, except for baselines with high entropy; temperature=1.0 and top-p=1.0}. We set the maximum length of the model generation to 512. This table corresponds to the Table~\ref{tab:table_main} version for LLaMA\update{2}, but only for watermark-based methods.}
\label{tab:llama_humaneval}
\vspace{4mm}
\end{table}

%% file: resources_acl/resources_appendix/fig_pareto_frontier_appendix.tex
\begin{figure*}[hbt!]
\begin{center}
\includegraphics[width=\textwidth]{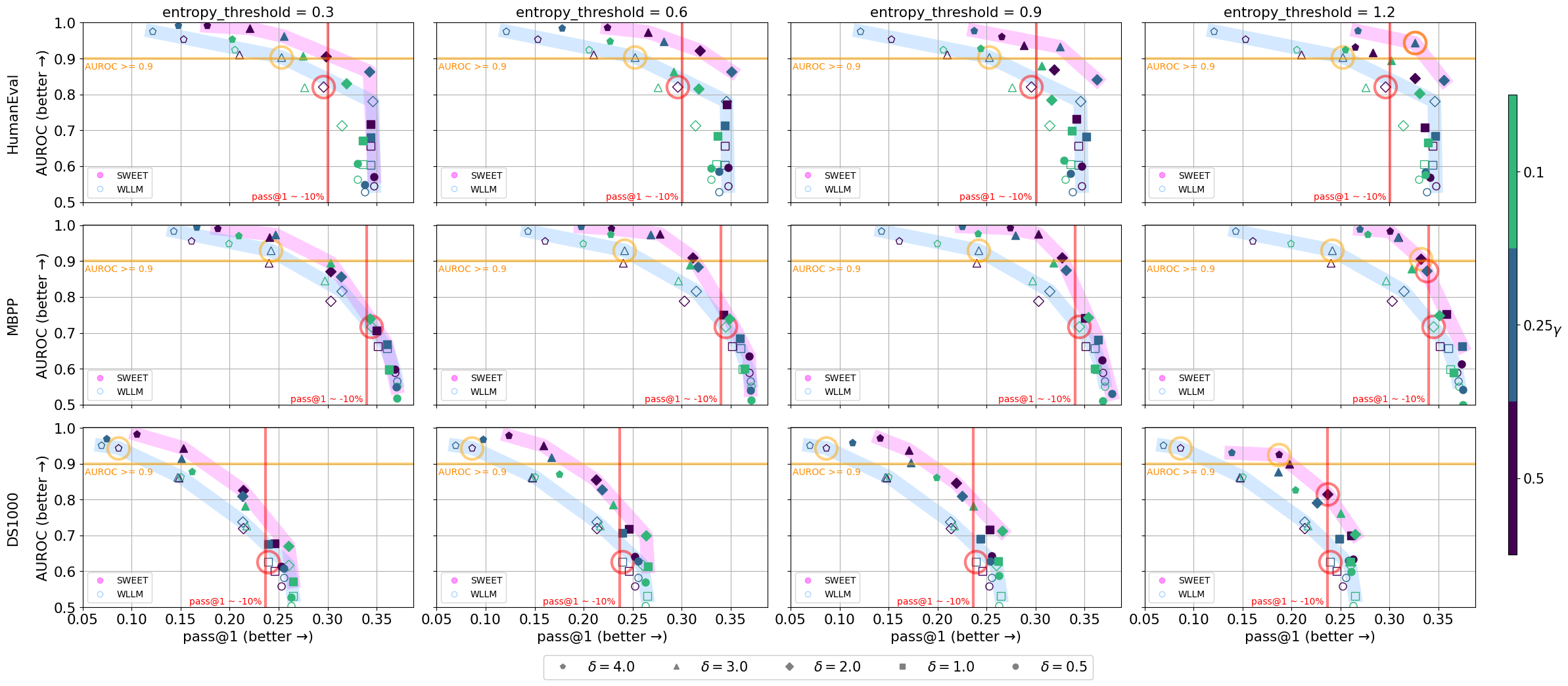}  
\end{center}
\caption{The tradeoff between AUROC and pass@1 of detecting real and generated samples of HumanEval, MBPP, and DS1000 datasets. The pink line represents a Pareto frontier of $\myalgo$, while the blue line represents that of $\wllm$. \update{In all tasks and the entropy threshold configurations,} $\myalgo$ shows consistent dominance. 
The \textcolor{red}{red}/\textcolor{orange}{orange} line and circles are the points used in Table~\ref{tab:table_main}.}
\label{fig:pareto_frontier_appendix}
\end{figure*}

%% file: resources_acl/resources_appendix/fig_pareto_frontier_cpp_java.tex
\begin{figure*}[hbt!]
\begin{center}
\includegraphics[width=\textwidth]{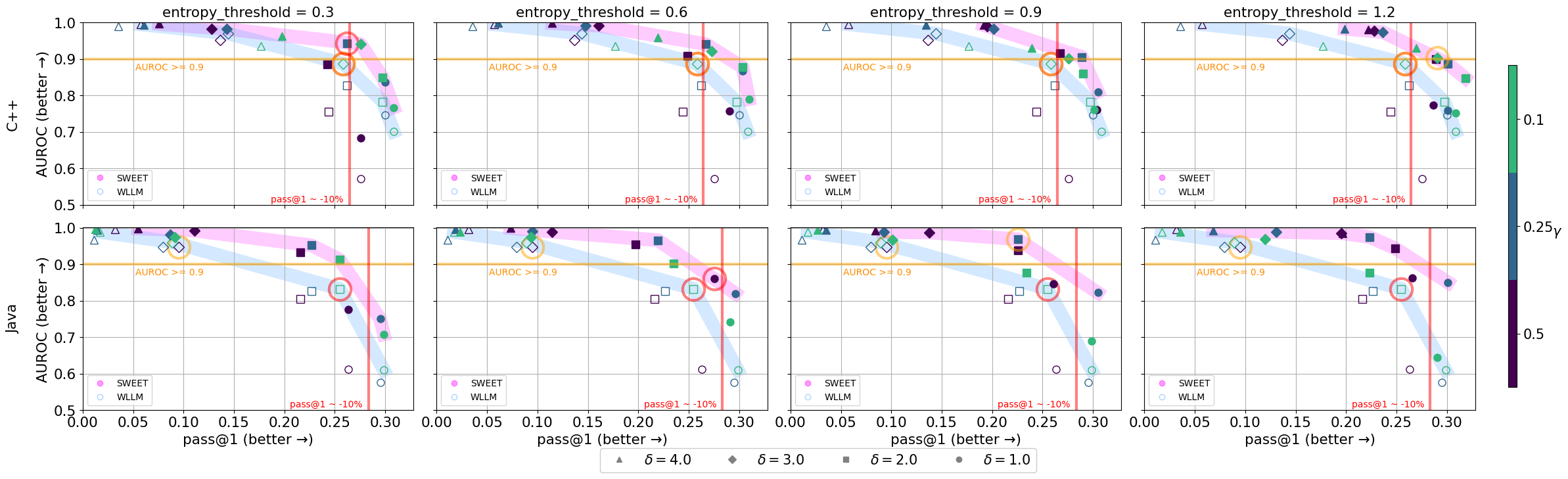}  
\end{center}
\caption{\cam{The tradeoff between AUROC and pass@1 of detecting real and generated samples of C++ and Java of HumanEvalPack datasets. The pink line represents a Pareto frontier of $\myalgo$, while the blue line represents that of $\wllm$. \update{In all tasks and the entropy threshold configurations,} $\myalgo$ shows consistent dominance. 
The \textcolor{red}{red}/\textcolor{orange}{orange} line and circles are the points used in Table~\ref{tab:table_main_2_w_classeval}.}}
\label{fig:pareto_frontier_cpp_java}
\end{figure*}

%% file: resources_acl/resources_appendix/fig_pareto_frontier_general_prompt.tex
\begin{figure*}[hbt!]
\begin{center}
\includegraphics[width=\linewidth]{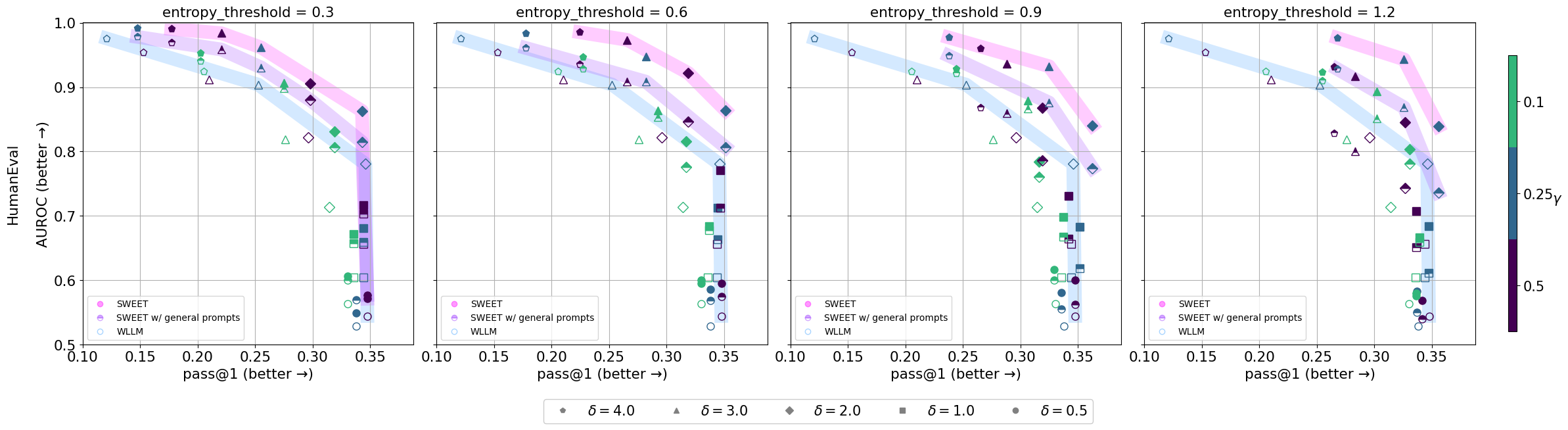}  
\end{center}
\caption{Effect of \textit{general prompts} in $\myalgo$ in HumanEval. In this setting, the detector does not know what information would have been included in a prompt if the given sample source code had been model-generated. $\myalgo$ appends the sample to the fixed number of `general prompts' that contain no information except for the format consistent with the answer.  The purple line represents the Pareto frontier of the `General prompts' version $\myalgo$. \update{Our approaches with general prompts still outperform $\wllm$ in both code quality preservation and watermark detection, drawing the Pareto frontiers ahead of those of $\wllm$.}}\label{fig:pareto_frontier_general_prefix_humaneval}
\end{figure*}

%% file: resources_acl/resources_appendix/fig_pareto_frontier_temp0.8.tex
\begin{figure*}[hbt!] 
\begin{center}
\includegraphics[width=\textwidth]{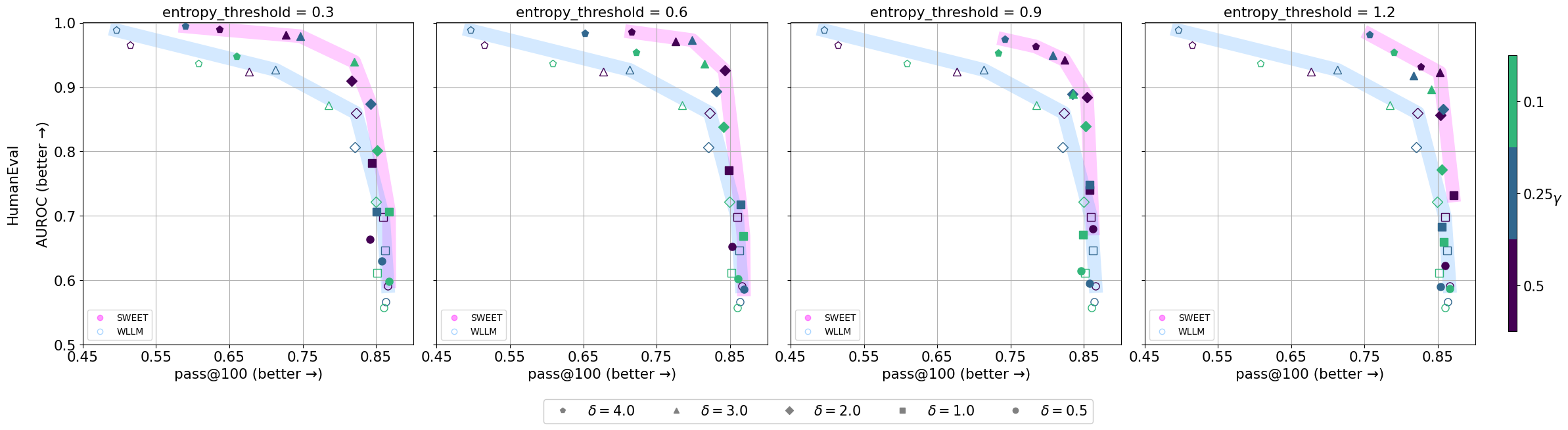}  
\end{center}
\caption{The tradeoff between AUROC and \textbf{pass@100} of detecting real and generated samples of HumanEval using temperature of 0.8 instead of 0.2 as other figures. We also generate $n=200$ outputs for calculating pass@100 scores. The pink line represents a Pareto frontier of $\myalgo$, while the blue line represents a Pareto frontier of $\wllm$. We observe consistent improvement in $\myalgo$.}\label{fig:pareto_frontier_humaneval_0.8_auroc}
\end{figure*}

%% file: resources_acl/resources_appendix/fig_pareto_frontier_llama2_and_surrogate.tex
\begin{figure*}[hbt!] 
\begin{center}
\includegraphics[width=\textwidth]{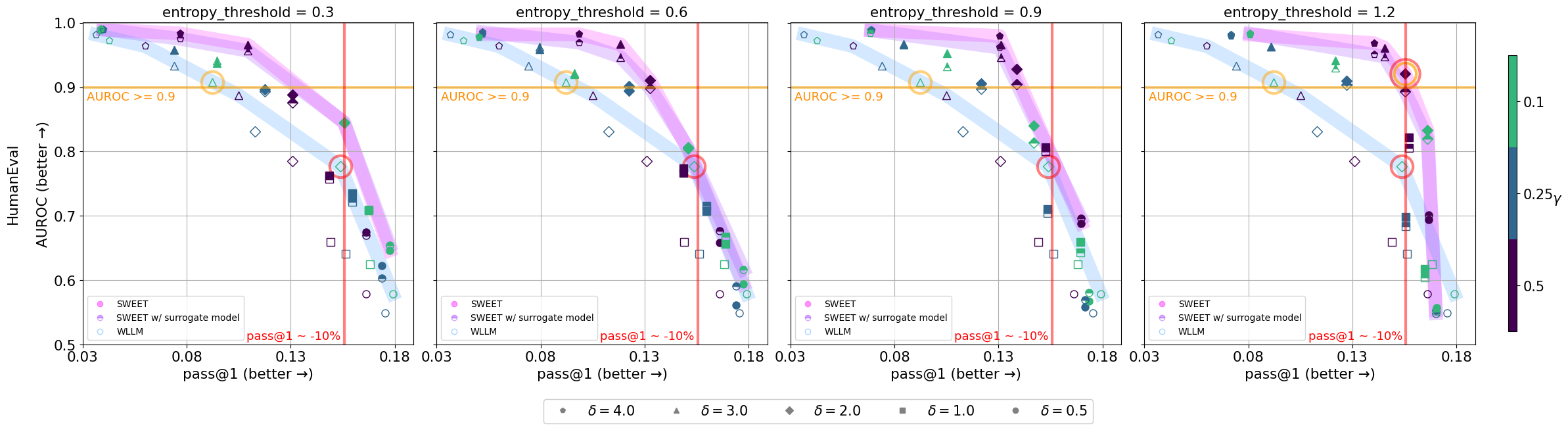}  
\end{center}
\caption{[LLaMa\update{2 13B} Results] The tradeoff between AUROC and pass@1 of detecting real and generated samples of HumanEval. The pink line represents a Pareto frontier of $\myalgo$, while the blue line represents a Pareto frontier of $\wllm$.
\update{Additionally, we include the results of the $\myalgo$ with the surrogate model (purple line), in which a smaller LM is used to detect watermarks to save computational costs. Our approaches mostly draw Pareto frontiers ahead of those of $\wllm$, even with the surrogate model.}
The \textcolor{red}{red}/\textcolor{orange}{orange} line and circles are the points used in Table~\ref{tab:llama_humaneval}.}
\label{fig:pareto_frontier_llama_humaneval}
\end{figure*}

%% file: resources_acl/resources_appendix/fig_detectability_appendix.tex
\begin{figure*}[hbt!]
\begin{center}
\includegraphics[width=\textwidth]{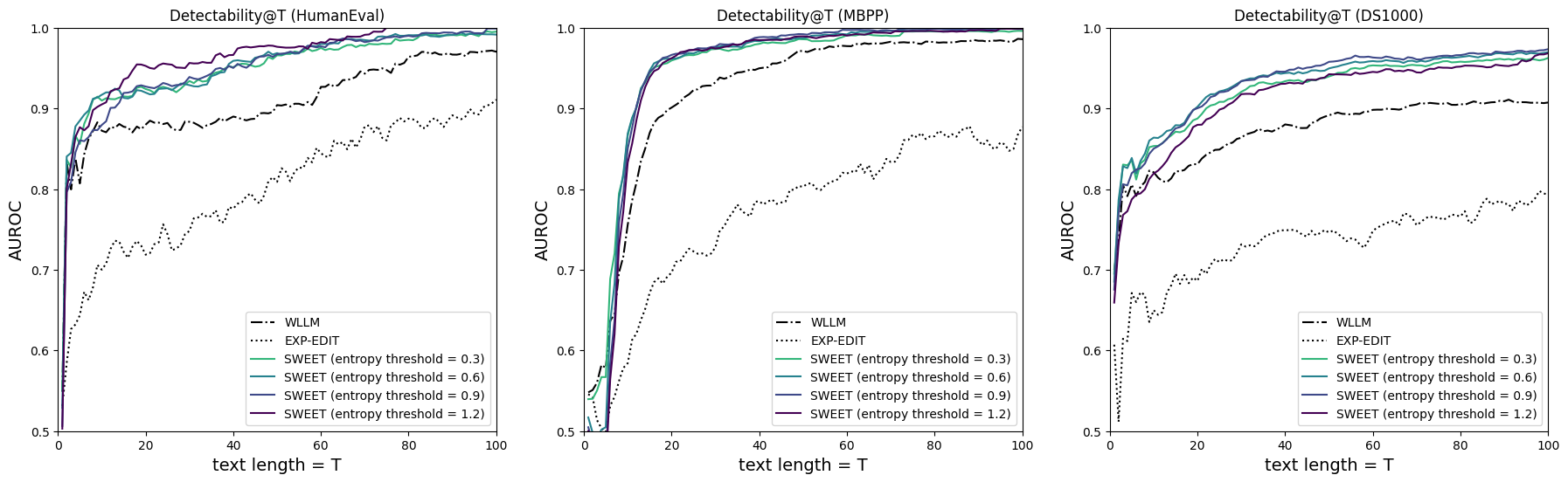}  
\end{center}
\caption{\update{Detectability@T~\citep{kirchenbauer2023reliability} at HumanEval, MBPP, and DS-1000. We set $\gamma=0.25$ and $\delta=3.0$ for $\wllm$ and $\myalgo$. For $\rdfw$, we use it with a high entropy setting. When calculating AUROC, we ensure at least 20 code texts of human-written solutions and machine-generated codes, respectively. We can observe that $\myalgo$ shows superior detection performance regardless of the text length in all tasks.}}
\label{fig:detectability_appendix}
\end{figure*}

%% file: resources_acl/resources_appendix/fig_entropy_calibration.tex
\begin{figure}[t!]
\begin{center}
\includegraphics[width=\linewidth]{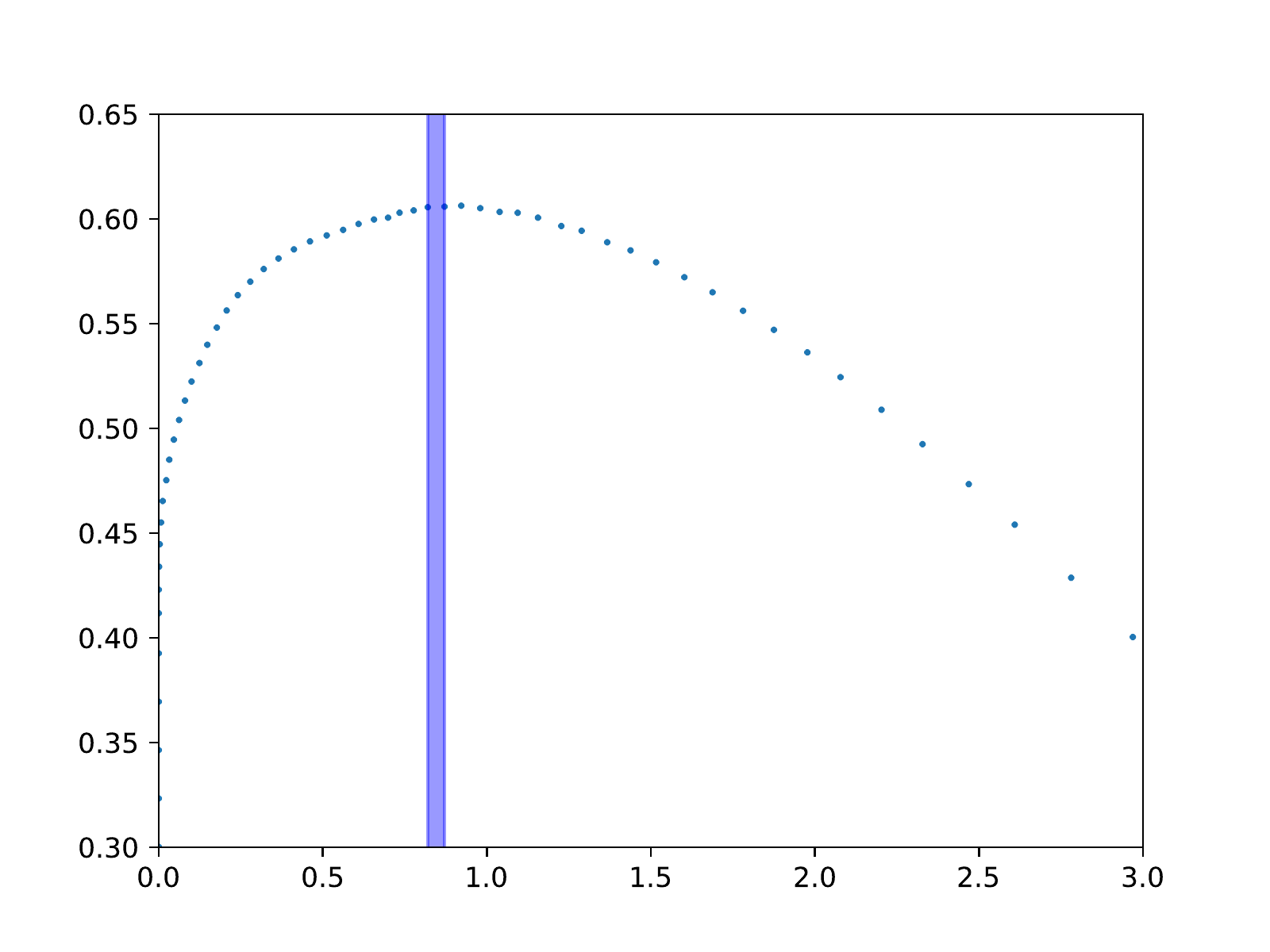}  
\end{center}
\caption{
\cam{The relationship between the entropy threshold and the pseudo-metric for z-score, $z'$ in Eq.~\ref{eq:z'}, calculated based on CodeSearchNet dataset. The blue region $[0.820, 0.871]$ is the best entropy threshold estimated by the calibration method described in Appendix~\ref{appendix:threshold_calibration}.
}}
\label{fig:threshold_calibration}
\end{figure}

%% file: resources_acl/resources_appendix/fig_lexical_distribution_above_entropy.tex
\begin{figure*}[t]
\begin{center}
\includegraphics[width=0.95\textwidth]{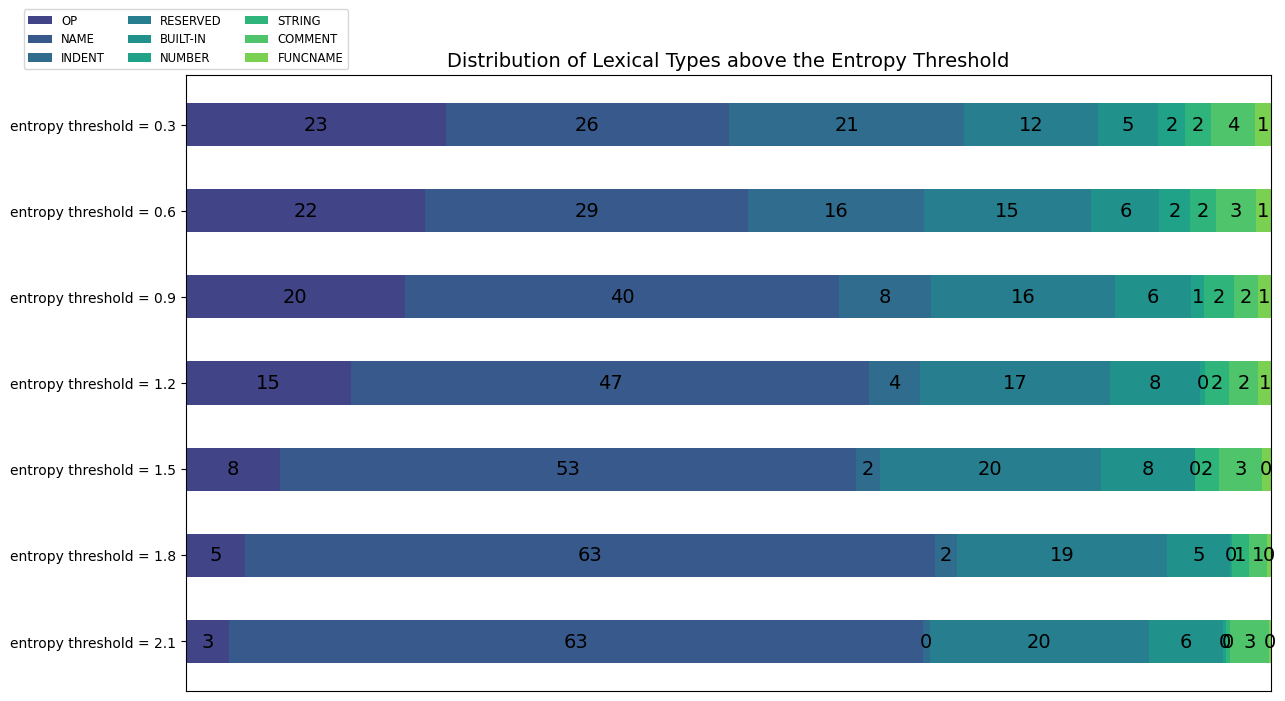}  
\end{center}
\caption{
Distribution of lexical types of $\myalgo$ output on HumanEval task. We draw examples when $\gamma=0.25$ and $\delta=3.0$. The proportion of NAME type tokens increases the most while that of INDENT type tokens converges to zero.
}
\label{fig:lex_dis_above_ent}
\end{figure*}

%% file: resources_acl/resources_appendix/fig_lexical_distribution_below_entropy.tex
\begin{figure*}[t]
\begin{center}
\includegraphics[width=0.95\textwidth]{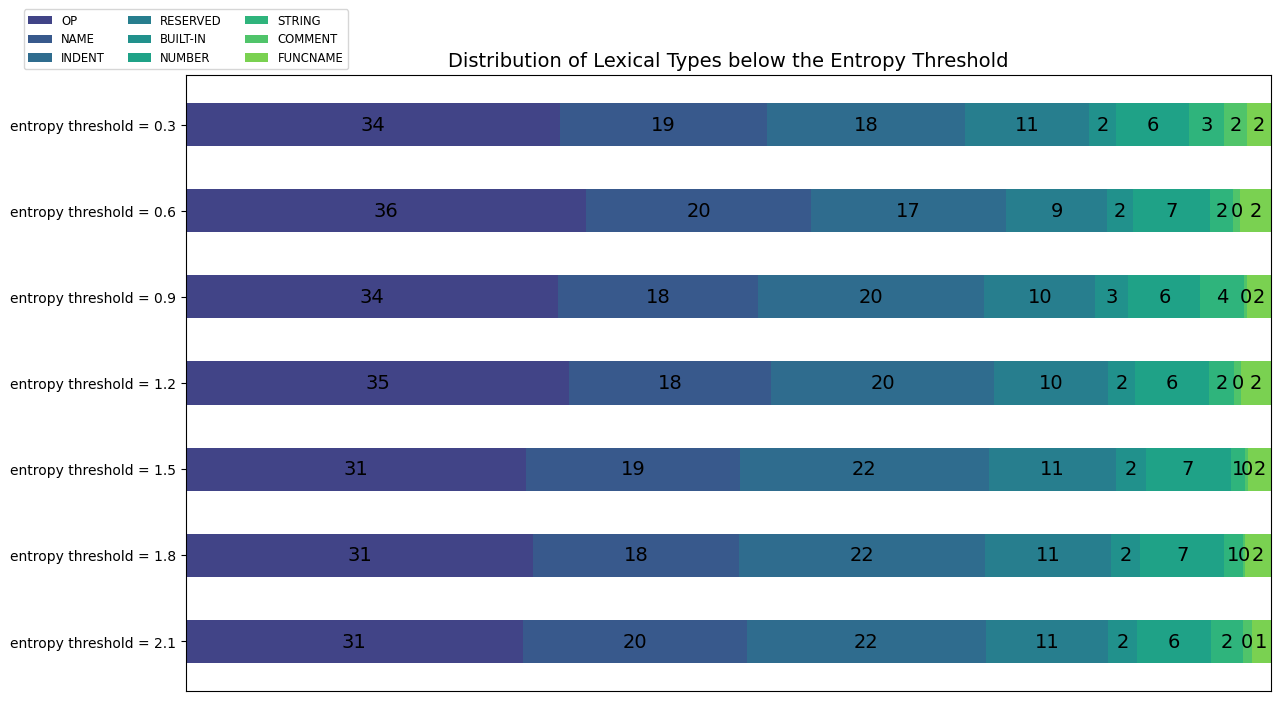}  
\end{center}
\caption{
Distribution of lexical types of $\myalgo$ output on HumanEval task. We draw examples when $\gamma=0.25$ and $\delta=3.0$. In contrast to the distributions above the threshold, there is almost no distribution change.
}
\label{fig:lex_dis_below_ent}
\end{figure*}